%% file: main.tex
\documentclass[10pt,twocolumn,letterpaper]{article}

\usepackage[pagenumbers]{iccv} 


\definecolor{iccvblue}{rgb}{0.21,0.49,0.74}
\usepackage[pagebackref,breaklinks,colorlinks,allcolors=iccvblue]{hyperref}

\def\paperID{4511} 
\def\confName{ICCV}
\def\confYear{2025}

\title{Direction-Aware Diagonal Autoregressive Image Generation}

\author{
\textbf{Yijia Xu}$^{1}$\quad
\textbf{Jianzhong Ju}$^{2}$\quad
\textbf{Jian Luan}$^{2}$\quad
\textbf{Jinshi Cui}$^{1*}$\quad \\
\textbf{\small $^*$corresponding author}\quad \vspace{2mm} \\
$^1$School of Intelligence Science and Technology, Peking University\quad \\
$^2$Xiaomi Inc.\quad \\
Codes and models:~\, \url{https://github.com/xiaomi-research/dar} \vspace{-7mm} \\
}

\usepackage{multirow}
\begin{document}
\twocolumn[{%
\renewcommand\twocolumn[1][]{#1}%
\maketitle
\begin{center}
    \centering
    \captionsetup{type=figure}
    \includegraphics[width=0.95\textwidth]{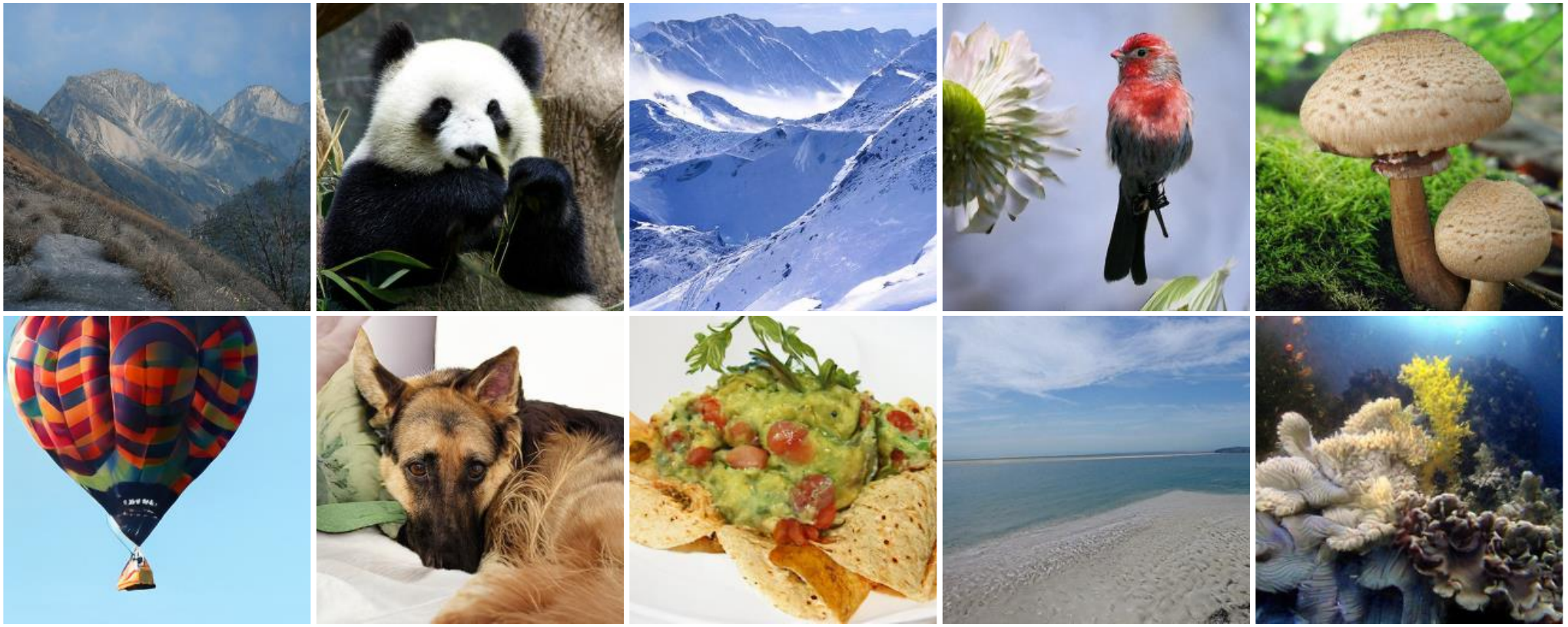}
    \captionof{figure}{\textbf{Generated samples of DAR.}
    We present samples generated by DAR, which is trained on the 256$\times$256 ImageNet dataset.}
    \label{fig:teaser}
\end{center}%
}]
\input{sec/0_abstract}
\input{sec/1_intro}
\input{sec/2_related}
\input{sec/3_method}

\input{sec/4_experiment}
\input{sec/5_conclusion}

\clearpage
{
    \small
    \bibliographystyle{ieeenat_fullname}
    \bibliography{main}
}

\input{sec/X_suppl}

\end{document}

%% file: sec/0_abstract.tex
\begin{abstract}
The raster-ordered image token sequence exhibits a significant Euclidean distance between index-adjacent tokens at line breaks, making it unsuitable for autoregressive generation.
To address this issue, this paper proposes Direction-Aware Diagonal Autoregressive Image Generation (DAR) method, which generates image tokens following a diagonal scanning order.
The proposed diagonal scanning order ensures that tokens with adjacent indices remain in close proximity while enabling causal attention to gather information from a broader range of directions.
Additionally, two direction-aware modules: 4D-RoPE and direction embeddings are introduced, enhancing the model's capability to handle frequent changes in generation direction.
To leverage the representational capacity of the image tokenizer, we use its codebook as the image token embeddings.
We propose models of varying scales, ranging from 485M to 2.0B.
On the 256$\times$256 ImageNet benchmark, our DAR-XL (2.0B) outperforms all previous autoregressive image generators, achieving a state-of-the-art FID score of 1.37.
\end{abstract}

%% file: sec/1_intro.tex
\section{Introduction}
\label{sec:intro}

In recent years, autoregressive models have been extensively utilized in various domains, including language generation, image generation, and multimodal generation.
The next-token prediction task has been a fundamental approach in large language models~\cite{llm_1,llm_2,llm_3,llm_4}, demonstrating remarkable capabilities in language generation tasks.
Concurrently, the application of next-token prediction in image generation tasks has made rapid advances~\cite{ar_9,ar_4,rar,llamagen,ibq}, revealing significant potential and promising results.
Furthermore, vanilla autoregressive approaches have indicated a promising pathway toward establishing a unified framework for multimodal generation~\cite{vlm_6,2drope_1,vlm_7,vlm_9,vlm_10,emu3}.

Unlike text sequences that inherently follow a unidirectional left-to-right ordering, discrete image token sequences produced by visual tokenizers maintain two-dimensional spatial coordinates.
The predominant raster scan ordering mechanism unwraps these image tokens row-wise, resulting in a large Euclidean distance between tokens with adjacent indices at line breaks.
Moreover, the prediction direction at line breaks significantly differs from that at other positions.

Several studies~\cite{rar,ar_4} have attempted various generation orders, yet none have surpassed the effectiveness of the raster scan order.
The underlying reason is that alternative generation orders introduce more complex directional variations, for which the model lacks adequate directional modeling capabilities.
For example, RAR~\cite{rar} incorporates absolute positional embeddings of next locations to train with random orders.
Although this allows the model to perceive the target positions, it fails to encode the generation directions and the random-order training is incompatible with unified multimodal generative models.
On the other hand, recent works~\cite{llamagen,ibq} have primarily focused on improving the reconstruction performance of image tokenizers, yet they fail to fully harness the representational capacity of these tokenizers when training autoregressive transformers.
To mitigate the limitations of next-token prediction, several studies~\cite{var,mar} have proposed variants of autoregressive methods. However, these approaches diverge significantly from conventional language models, making them challenging to integrate into a unified multimodal architecture.

\begin{figure}[t]
    \centering
    \begin{subfigure}{0.47\linewidth}
        \includegraphics[width=\linewidth]{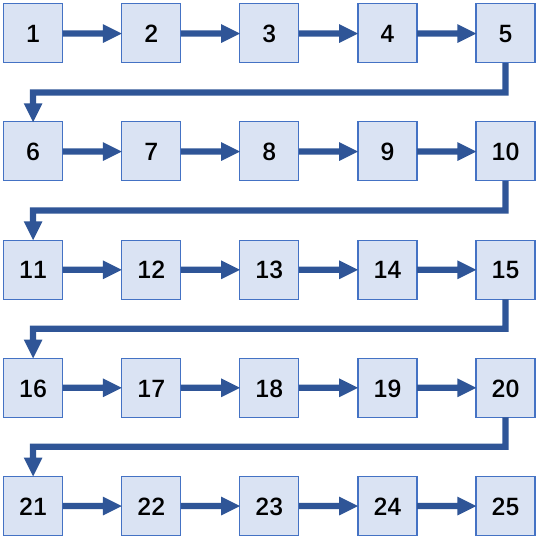}
        \caption{Raster scan order.}
        \label{fig:order_raster}        
    \end{subfigure}
    \hfill
    \begin{subfigure}{0.47\linewidth}
        \includegraphics[width=\linewidth]{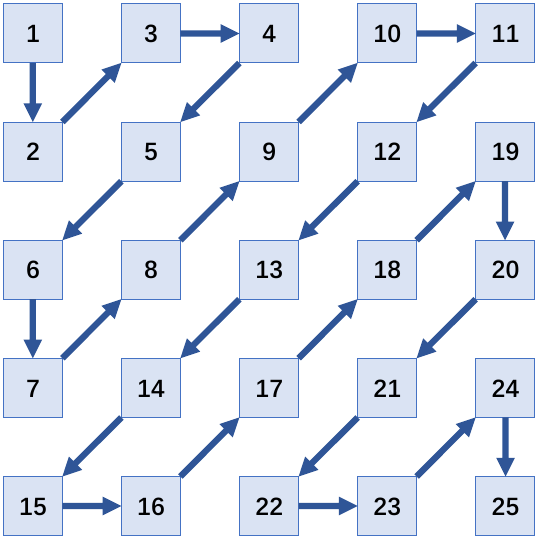}
        \caption{Diagonal scan order.}
        \label{fig:order_diag}        
    \end{subfigure}
    \caption{\textbf{Illustration of two image token arrangements.} Subfigure \subref{fig:order_raster} shows the raster scan order. Subfigure \subref{fig:order_diag} shows the diagonal scan order.}
    \label{fig:order}
\end{figure}

To address these challenges, we propose a novel autoregressive modeling method, namely, Direction-Aware Diagonal Autoregressive Image Gnenration (DAR).
Specifically, as shown in \cref{fig:order_diag}, we rearrange the image tokens in diagonal scanning order, ensuring that all tokens with adjacent indices are positioned in close proximity.
Since the tokens are distributed along the diagonal connecting the bottom-left to the top-right, this ordering allows the causal attention mechanism to capture information from a broader range of directions compared to the traditional raster scan order.
To enable the model to adapt to frequently changing generation directions, we propose 4D-RoPE that incorporates both the current position and the next position.
The variations in relative positions between any two tokens are injected into the attention matrix, enabling the model to effectively handle sequences with frequently varying directions.
Furthermore, we use direction embeddings to directly represent generation directions and utilize them to calculate the scale and shift parameters in AdaLN~\cite{adaln}.
To fully leverage the representation capabilities of the image tokenizer, we utilize its codebook as image token embeddings and freeze these parameters.
Notably, our proposed method strictly adheres to the ``next-token prediction'' paradigm, which can be straightforwardly extended to a multimodal framework.

We summarize our contributions as follows:
\begin{itemize}
    \item[$\bullet$] We introduce a diagonal scanning order for autoregressive image generation, which enables causal attention to capture information from more directions while maintaining close proximity between index-adjacent tokens.
    \item[$\bullet$] We introduce two direction-aware modules: 4D-RoPE and direction embeddings.
    The 4D-RoPE enables each token's position embedding to incorporate its own position and the position of its next token.
    Direction embeddings are used to calculate parameters in AdaLN.
    These modules enhance the model's capability to handle frequently changing generation directions.
    \item[$\bullet$] We introduce a family of autoregressive image generation models, ranging in size from 485M to 2.0B.
    Our DAR-XL (2.0B) achieves an FID score of 1.37, surpassing all other autoregressive approaches and establishing state-of-the-art performance on the 256$\times$256 ImageNet benchmark~\cite{imagenet}.
\end{itemize}




%% file: sec/2_related.tex
\section{Related Work}
\label{sec:related}

\subsection{Visual Generation Models.}
In the field of visual generation, there exists a diverse array of models. 
Generative Adversarial Networks (GANs)~\cite{GAN,GAN_2,Karras_2019_CVPR} introduce a novel adversarial training mechanism. 
Diffusion models~\cite{Diff_1,Diff_2,Diff_3,Diff_4,adaln,Diff_6,Diff_7} generate high-quality samples by gradually denoising data through a Markov chain process. 
Masked generative models~\cite{mask_1,mask_2,mask_3,mask_4,mask_5} use a BERT-style approach to predict masked tokens.
Pioneering research in autoregressive visual modeling~\cite{ar_1,ar_2,ar_3} leverages the architecture of language models to generate images by predicting pixels.
By utilizing image tokenizers, autoregressive models shift to producing discrete token sequences in a predict next-token manner~\cite{ar_4,ar_5,ar_6,ar_7,ar_8,rar,llamagen,ibq}.
These approaches generate image tokens in a raster scan order, which fails to adequately exploit the inherent 2D structural characteristics of images.
VAR~\cite{var} introduces a generative approach for next-scale prediction and MAR~\cite{mar} employs diffusion loss for the training of masked autoregressive models.
These variants of autoregressive methods fundamentally differ from the training objectives of language models, making them suboptimal for building multimodal generative frameworks.
In this work, we maintain the approach of predicting next-token, utilizing a diagonal scanning order to generate image tokens and enhance the generation quality with direction-aware modules.

\subsection{Position Embedding in Vision Modeling}
Absolute Positional Embedding (APE)~\cite{ape_1,ape_2} employs learnable parameters to represent positional information.
Relative Position Bias (RPB)~\cite{rpb_1,rpb_2,rpb_3} focuses on the relative distances between tokens.
RoFormer~\cite{1drope} proposes Rotary Position Embedding (RoPE), which injects relative positions to the attention matrix in rotation form.
2D-RoPE applies 1D-RoPE separately to each of the two dimensions, making it more suitable for image encoding. This approach has been utilized in Vision Transformers (ViTs)~\cite{eva02,2drope_1} and diffusion models~\cite{2drope_2}.
Mixed frequency 2D-RoPE~\cite{mixrope} enables diagonal direction handling and makes 2D-RoPE learnable.

\subsection{Multimodal Foundation Models}
Vision language models~\cite{vlm_1,vlm_2,vlm_3,vlm_4,vlm_5} have demonstrated remarkable capabilities in visual understanding through the integration of alignment training and instruction tuning.
Pioneering methods~\cite{vlm_6,2drope_1,vlm_7,vlm_9,vlm_10} have demonstrated the potential of training multimodal models with unified next-token prediction objectives, enabling simultaneous understanding and generation capabilities.
However, such models remain in the early stages of research.
In this work, we focus on enhancing the next-token prediction method to better suit image generation tasks, while ensuring that our approach remains applicable for training multimodal models.

%% file: sec/3_method.tex
\section{Method}
\label{sec:method}

\begin{figure*}[t]
    \centering
    \includegraphics[width=\linewidth]{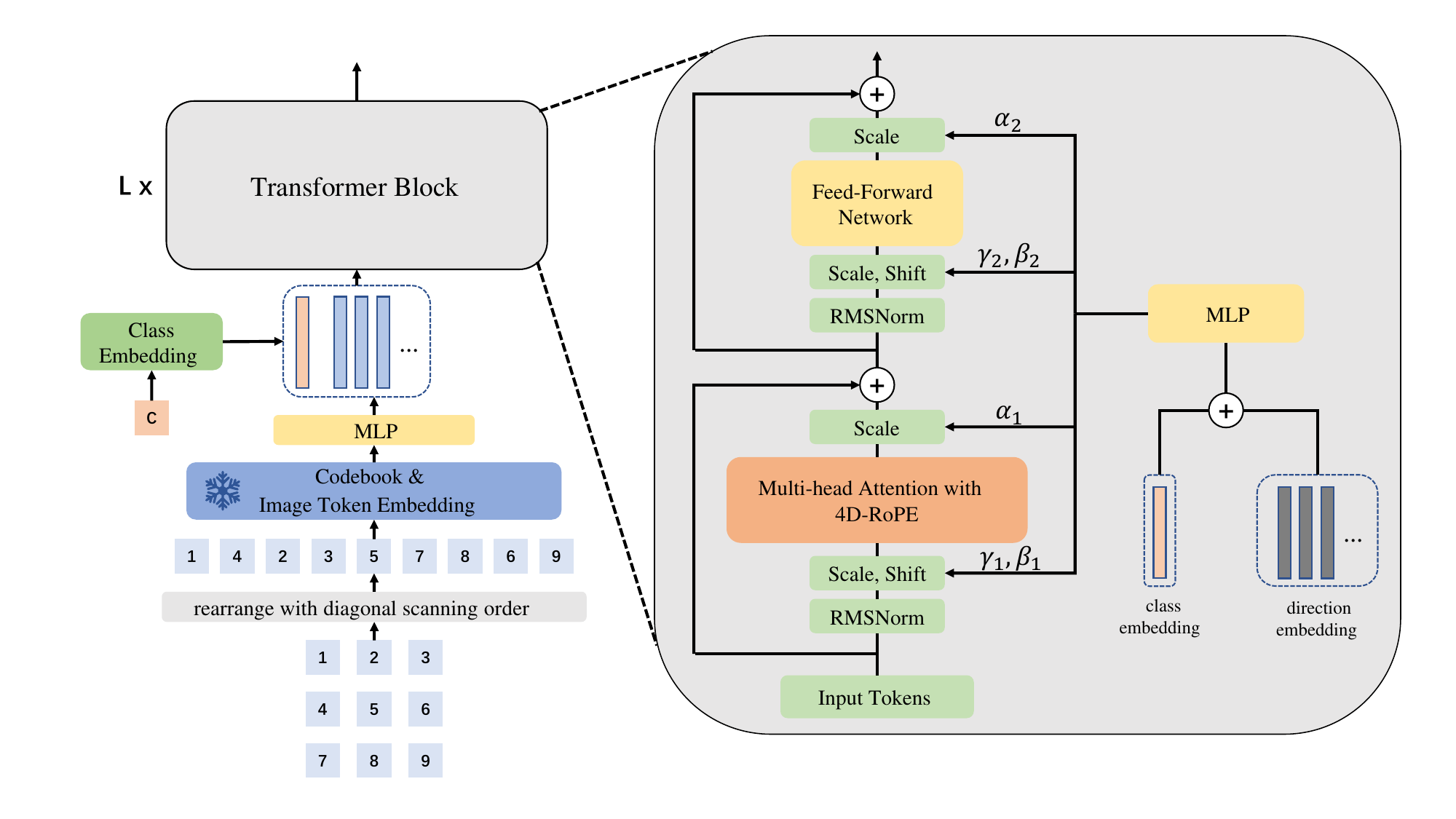}
    \caption{\textbf{Overview of the proposed Direction-Aware Diagonal Autoregressive (DAR) model.}
    The discrete image tokens are arranged in the diagonal scanning order.
    These tokens are then processed through the codebook within the image tokenizer and MLP to obtain image token embeddings.
    The class embedding is prepended to the sequence, which is subsequently fed into the autoregressive transformer.
    Within the transformer block, 4D-rope that combines both the current and next positions is employed during multi-head attention.
    AdaLN calculates the scale and shift parameters using the sum of class embeddings and direction embeddings.
    }
    \label{fig:model}
\end{figure*}

\Cref{fig:model} provides an overview of our proposed Direction-Aware Diagonal Autoregressive (DAR) model.
In \cref{pre}, we provide an overview of autoregressive image generation.
\cref{subsec:3.2} introduces the diagonal scanning order, while \cref{subsec:4drope} proposes the 4D-RoPE method and direction embeddings for optimizing multi-directional generation.
Finally, \cref{subsec:3.4} presents an approach for effectively utilizing the image tokenizer to obtain image token embeddings.

\subsection{Preliminary}
\label{pre}
The framework for autoregressive image generation comprises a vector-quantized image tokenizer and an autoregressive transformer.

The image tokenizer adopts an encoder-quantizer-decoder architecture, in which the quantizer utilizes a codebook $\mathcal{C} \in \mathbb{R}^{K \times D}$, where $K$ denotes the codebook size and $D$ denotes the code dimension.
An input image $\mathcal{I} \in \mathbb{R}^{H \times W \times 3}$ is first encoded by the encoder into the feature map $\mathcal{Z} \in \mathbb{R}^{h \times w \times D}$, where $h=H/p$, $w=W/p$, $p$ is the downsample ratio.
The feature map is subsequently quantized into discrete tokens $\mathcal{Q} \in \mathbb{R}^{h \times w \times D}$ via the codebook $\mathcal{C}$.
$\mathcal{Q}$ is flattened in raster order to obtain the discrete token sequence $\mathbf{x} = [x_1,x_2,...,x_T]$, where $T=h \times w$.
The autoregressive transformer is trained on these sequences using the next-token prediction task, thus acquiring the capability for autoregressive image generation.
Specifically, the optimization objective is to maximize the following likelihood:
\begin{equation}
    p(\mathbf{x}|c) = \prod_{t=1}^{T}p(x_t|x_1,x_2,...,x_{t-1},c)
\end{equation}
, where c is the condition.
During inference, the predicted token sequence is transformed into image pixels by the decoder of the image tokenizer.

For position embedding in the transformer, previous methods employ 2D-RoPE to inject 2D positional information of image tokens into the attention matrix~\cite{llamagen,ibq}.
Specifically, the frequencies utilized in 2D-RoPE are calculated as:
\begin{equation}
    \theta_t = 10000^{-t/(d_{head}/4)}
\end{equation}
, where $d_{head}$ denotes attention head dimension, $t \in \{0,1,...,d_{head}/4\}$.
For a 2D position $\mathbf{p}_n=(p_n^x,p_n^y)$ where $p_n^x \in \{0,1,...,h\}$, $p_n^y \in \{0,1,...,w\}$, the rotation matrix $\mathbf{R} \in \mathbb{C}^{T \times (d_{head}/2)}$ is defined as
\begin{equation}
\label{2drot}
    \mathbf{R}(n,2t)=e^{i \theta_t p_n^x}, \mathbf{R}(n,2t+1)=e^{i \theta_t p_n^y}
\end{equation}
, where $t \in \{0,1,...,d_{head}/4\}$.
To inject positional information, queries and keys $\mathbf{q},\mathbf{k} \in \mathbb{R}^{T \times d_{head}}$ are transformed into complex matrices $\bar{\mathbf{q}},\bar{\mathbf{k}} \in \mathbb{C}^{T \times (d_{head}/2)}$. 
Then, $\bar{\mathbf{q}}$ and $\bar{\mathbf{k}}$ are element-wise multiplied with the rotation matrix $\mathbf{R}$:
\begin{equation}
\label{applyrot}
    \bar{\mathbf{q}}^{\prime} = \bar{\mathbf{q}} \odot \mathbf{R},
    \bar{\mathbf{k}}^{\prime} = \bar{\mathbf{k}} \odot \mathbf{R}
\end{equation}.
Using $\bar{\mathbf{q}}^{\prime}$ and $\bar{\mathbf{k}}^{\prime}$ for subsequent calculations, the model can effectively incorporate 2D relative positional information.

For class-conditional generation, previous methods~\cite{rar,ibq} concatenate the class token at the beginning of the sequence and employ adaptive layer normalization (AdaLN)~\cite{adaln} within the autoregressive transformer.
They calculate the scale and shift parameters in AdaLN using the class token alone or the sum of the class token and timestep embeddings.

Directly applying the next-token prediction on a raster-ordered image token sequence is suboptimal.
Specifically, the embedding of the last token in each row, i.e. the token at position $(i, w)$, after passing through multiple transformer blocks, is used to predict the probability distribution of the first token in the next row.
During the modeling process, the current token consistently utilizes the position embedding of $(i, w)$, which is significantly distant from the token to be predicted at position $(i+1, 1)$.

\subsection{Diagonal Scanning Order}
\label{subsec:3.2}
In the raster order sequence, a number of tokens are positioned far from their subsequent tokens, resulting in inaccurate position embeddings during the modeling process.
Specifically, for a token $x_{cur}$ located at position $\mathbf{p}_{cur}$, we refer to its next token as $x_{nxt}$, which is positioned at $\mathbf{p}_{nxt}$.
When the embedding of $x_{cur}$ is initially fed into the transformer, it is appropriate to use $\mathbf{p}_{cur}$ for position embedding.
The hidden states output by the final layer are utilized to calculate the probability distribution of $x_{nxt}$.
$\mathbf{p}_{nxt}$ represents the most accurate position of them.
Consequently, the position embeddings most suitable for the hidden states in the intermediate layers lie between $\mathbf{p}_{cur}$ and $\mathbf{p}_{nxt}$.
Since RoPE is a relative position embedding, it is accurate to use $\mathbf{p}_{cur}$ for position embeddings in each layer if $\mathbf{p}_{cur}$ and $\mathbf{p}_{nxt}$ are close, or if the relative positions of $\mathbf{p}_{cur}$ and $\mathbf{p}_{nxt}$ for all tokens are equal.
However, in the raster scan order, $\mathbf{p}_{cur}$ and $\mathbf{p}_{nxt}$ are significantly distant at line break positions and their relative distance exhibits a substantial deviation compared to other positions.

To address these challenges, as shown in \cref{fig:order}, we propose a diagonal scanning order that ensures adjacent tokens are positioned in close proximity.
We initiate the generation process from the top-left corner, proceeding along a 45-degree diagonal direction. 
The generation alternates between two patterns: from the bottom-left to the top-right and from the top-right to the bottom-left.
The proximity between $\mathbf{p}_{cur}$ and $\mathbf{p}_{nxt}$ enhances the accuracy of position embeddings in the autoregressive transformer.
In addition, the diagonal scanning order offers another advantage: image tokens can acquire information from a more diverse range of directions.
Under the constraint of causal attention, $x_{cur}$ can only attend to the preceding tokens when predicting $x_{nxt}$.
Consequently, in the raster scan order, $x_{cur}$ is restricted to accessing information from the top-left and top-right regions only.
In contrast, when utilizing the diagonal scan order, $x_{cur}$ is able to gather information from three directions: the top-left, top-right, and bottom-left regions relative to $\mathbf{p}_{cur}$.
A broader field of view enhances the model's generative capabilities.

The diagonal scanning sequence introduces an additional issue, where the generation direction frequently changes.
There are a total of four generation directions: rightward, downward, upper-right, and lower-left.
In \cref{subsec:4drope}, we propose a method to address this problem.

\subsection{4D-RoPE and Direction Embeddings}
\label{subsec:4drope}

Since language sequences are generated unidirectionally from left to right, the prevalent next-token prediction paradigm in language models lacks explicit consideration of generation directionality.
When generating image tokens, there are two possible directions for raster order and four possible directions for diagonal scanning order.
Frequent changes in the generation direction significantly increase the difficulty of image generation.
To explicitly condition the model on the generation direction, we propose 4D-RoPE that combines $\mathbf{p}_{cur}$ and $\mathbf{p}_{nxt}$, and modify AdaLN to incorporate direction embeddings.

We redefine the position of the token with index $n$ as $\mathbf{p}_n=(p_n^x,p_n^y,p_{n+1}^x,p_{n+1}^y)$.
The rotation matrix $\mathbf{R} \in \mathbb{C}^{T \times (d_{head}/2)}$ in ~\cref{2drot} is changed as
\begin{gather}
    \mathbf{R}(n,4t)=e^{i \theta_t p_n^x}, \mathbf{R}(n,4t+1)=e^{i \theta_t p_n^y}, \nonumber \\
    \mathbf{R}(n,4t+2)=e^{i \theta_t p_{n+1}^x}, \mathbf{R}(n,4t+3)=e^{i \theta_t p_{n+1}^y}.
\end{gather}
Also, we make corresponding modifications to frequencies
\begin{equation}
    \theta_t = 10000^{-t/(d_{head}/8)}
\end{equation}
, where $t \in \{0,1,...,d_{head}/8\}$.
In this way, the calculated $\bar{\mathbf{q}}^{\prime}$ and $\bar{\mathbf{k}}^{\prime}$ from \cref{applyrot} simultaneously incorporate the positional information of both the current token and the next token.
We refer to this method of using four-dimensional coordinates for RoPE as 4D-RoPE.
After injecting the information of both $\mathbf{p}_{cur}$ and $\mathbf{p}_{nxt}$ into the attention matrix, the model can effectively distinguish between different generation directions.
As discussed in \cref{pre}, the precise positions of hidden states in each layer lie between $\mathbf{p}_{cur}$ and $\mathbf{p}_{nxt}$.
When multiple generation directions exist, the relative positions of any two tokens may change across layers.
4D-RoPE enables the model to perceive changes in the relative positions, thereby enhancing its ability to adapt to variations in generation directions.

To further enhance the model's capability to handle diverse generation directions, we introduce direction embeddings and incorporate them into AdaLN.
Specifically, when using diagonal scanning order, we utilize four learnable embeddings to represent four distinct generation directions.
Each index selects the corresponding direction embedding.
The direction embeddings are then summed with the class embedding to calculate the scale and shift parameters in AdaLN.
Our enhanced AdaLN condition on directional information, thereby improving the model's capability to generate sequences with frequent directional variations.

\subsection{Codebook-based Image Token Embeddings}
\label{subsec:3.4}

After pre-training, the image tokenizer exhibits robust capabilities in image representation. To fully leverage the potential of the image tokenizer, unlike previous methods~\cite{rar,llamagen} that learn image token embeddings from scratch, we directly utilize the codebook from the image tokenizer as the image token embeddings and freeze their parameters.
Subsequently, we use an MLP to align the dimension of the embeddings with the input dimension of the transformer.
By leveraging the codebook designed for image generation tasks, we can more effectively train the autoregressive transformer to predict the next token.


%% file: sec/4_experiment.tex
\section{Experiments}
\label{sec:experiment}

\subsection{Implementations Details}
\label{subsec:imp}
Following previous works~\cite{llamagen,ibq,rar}, our implementation of DAR consists of two main modules: an image tokenizer and an autoregressive transformer.

\noindent
\textbf{Vector Quantized Image Tokenizer.}
We use the IBQ~\cite{ibq} with the official weight, trained on 256$\times$256 ImageNet~\cite{imagenet}, as our image tokenizer.
It achieves a rFID~\cite{rfid} of 1.37, with a codebook size 16,384 and a code dimension 256.
Following the same architecture of VQGAN~\cite{ar_4}, it converts 256$\times$256 resolution images into 16$\times$16 discrete tokens.

\noindent
\textbf{Autoregressive Transformer.}
We adopt a Llama-based architecture~\cite{llm_1,llm_2}, integrating SwiGLU~\cite{swiglu} and RMSNorm~\cite{rmsnorm} within the transformer, and enhance the RoPE mechanism to a 4D-RoPE that combines both the current and the next positions.
Furthermore, we utilize AdaLN~\cite{adaln} with direction embeddings to improve the image generation capabilities of the model.
Following previous works~\cite{llamagen,modelcfg_1,llm_1}, we adopt three model configurations: DAR-B, DAR-L, and DAR-XL.
Detailed configurations and model sizes are shown in \cref{tab:modelconfig}
\begin{table}
  \centering
  \resizebox{\linewidth}{!}{
  \begin{tabular}{l|cccc}
    \toprule
    Model & Layers & Hidden Size & Heads & \#Para. \\
    \midrule
    DAR-B & 24 & 1024 & 16 & 485M \\
    DAR-L & 36 & 1280 & 20 & 1117M \\
    DAR-XL & 48 & 1536 & 24 & 2077M \\
    \bottomrule
  \end{tabular}
  }
  \caption{\textbf{Architecture configurations and model sizes of DAR.} We follow previous works~\cite{llamagen,modelcfg_1,llm_1} for configurations.}
  \label{tab:modelconfig}
\end{table}

\noindent
\textbf{Image Token Embeddings.}
We use the codebook from the image tokenizer as image token embeddings, and freeze these parameters during training.
Subsequently, we utilize a two-layer MLP to align the 256-dimensional embeddings with the input dimension of the transformer.

\noindent
\textbf{Dataset.}
We train our model on the 256$\times$256 ImageNet-1K~\cite{imagenet} dataset, which comprises 1,000 classes and a total of 1,281,167 images.
Following previous works~\cite{llamagen,rar}, we use the ten-crop data augmentation technique to enhance the diversity of the training set.
Additionally, we utilize IBQ~\cite{ibq} to pre-tokenize image codes, thereby improving training efficiency.

\noindent
\textbf{Training Protocols.}
We train the model for 400 epochs with batch size 2048.
The warm-up strategy is employed for the initial 100 epochs, during which the learning rates for DAR-B are linearly increased from $0$ to $1 \times 10^{-3}$, while for DAR-L and DAR-XL, the learning rate is increased to $4 \times 10^{-4}$.
For the subsequent 300 epochs, a cosine decay schedule is applied to gradually reduce the learning rates of all models to 1e-5.
All models are trained with AdamW~\cite{adamw_1, adamw_2} optimizer with $\beta_1 = 0.9$, $\beta_2 = 0.96$, weight decay $= 0.05$, gradient clipping of 1.0.
A dropout rate of 0.1 is utilized across the token embedding, attention modules, and feed-forward network modules.
Additionally, a class condition dropout of 0.1 is implemented during the training process.

\noindent
\textbf{Evaluation metrics.}
The evaluation metrics are Fréchet Inception Distance (FID)~\cite{rfid}, Inception Score (IS)~\cite{IS}, and Precision/Recall~\cite{prerec}.
All metrics are computed using the scripts provided by~\cite{Diff_1} for fair comparisons.

\noindent
\textbf{Sampling Protocols.}
We sample 50,000 images to compute evaluation metrics. Following previous works~\cite{rar}, we employ classifier-free guidance~\cite{cfg} and the power-cosine guidance schedule~\cite{cosinecfg}, without utilizing top-k or top-p techniques.
The detailed hyper-parameters for each model can be found in appendix.

\subsection{Main Results}
\label{subsec:mainresult}

\begin{table}
  \centering
  \resizebox{\linewidth}{!}{
  \begin{tabular}{c l c c c c c}
    \toprule
    Type & Model & \#Para. & FID$\downarrow$ & IS$\uparrow$ & Pre.$\uparrow$ & Rec.$\uparrow$ \\
    \midrule 
    \multirow{3}{*}{Diffusion} & LDM-4 \cite{Diff_6} & 400M & 3.60 & 247.7 & 0.87 & 0.48 \\
    & DiT-XL/2 \cite{adaln} & 675M & 2.27 & 278.2 & 0.83 & 0.57 \\
    & MDTv2-XL/2 \cite{cosinecfg} & 676M & 1.58 & 314.7 & 0.79 & 0.65 \\
    \midrule
    \multirow{4}{*}{Mask.} & MaskGIT \cite{mask_1} & 177M & 6.18 & 182.1 & - & - \\
    & TiTok-S-128 \cite{titok} & 287M & 1.97 & 281.8 & - & - \\
    & MAGVIT-v2 \cite{mask_5} & 307M & 1.78 & 319.4 & - & - \\
    & MaskBit \cite{mask_3} & 305M & 1.52 & 328.6 & - & - \\
    \midrule
    \multirow{2}{*}{VAR} & VAR-d30 \cite{var} & 2.0B & 1.92 & 323.1 & 0.82 & 0.59 \\
    & VAR-d30-re \cite{var} & 2.0B & 1.73 & 350.2 & 0.82 & 0.60 \\
    \midrule
    \multirow{3}{*}{MAR} & MAR-B \cite{mar} & 208M & 2.31 & 281.7 & 0.82 & 0.57 \\
    & MAR-L \cite{mar} & 479M & 1.78 & 296.0 & 0.81 & 0.60 \\
    & MAR-H \cite{mar} & 943M & 1.55 & 303.7 & 0.81 & 0.62 \\
    \midrule
    \multirow{4}{*}{FlowAR} & FlowAR-S \cite{flowar} & 170M & 3.61 & 234.1 & 0.83 & 0.50 \\
    & FlowAR-B \cite{flowar} & 300M & 2.90 & 272.5 & 0.84 & 0.54 \\
    & FlowAR-L \cite{flowar} & 589M & 1.90 & 281.4 & 0.83 & 0.57 \\
    & FlowAR-H \cite{flowar} & 1.9B & 1.65 & 296.5 & 0.83 & 0.60 \\
    \midrule
    \multirow{17}{*}{AR} & VQGAN \cite{ar_4} & 1.4B & 15.78 & 74.3 & - & - \\
    & VQGAN-re \cite{ar_4} & 1.4B & 5.20 & 280.3 & - & - \\
    & ViT-VQGAN \cite{ar_8} & 1.7B & 4.17 & 175.1 & - & - \\
    & ViT-VQGAN-re \cite{ar_8} & 1.7B & 3.04 & 227.4 & - & - \\
    & MonoFormer \cite{mono} & 1.1B & 2.57 & 272.6 & 0.84 & 0.56 \\
    & Open-MAGVIT2-XL \cite{MAG} & 1.5B & 2.33 & 271.8 & 0.84 & 0.54 \\
    & LlamaGen-XL-384 \cite{llamagen} & 775M & 2.62 & 244.1 & 0.80 & 0.57 \\
    & LlamaGen-XXL-384 \cite{llamagen} & 1.4B & 2.34 & 253.9 & 0.80 & 0.59 \\
    & LlamaGen-3B-384 \cite{llamagen} & 3.1B & 2.18 & 263.3 & 0.81 & 0.58 \\
    & IBQ-B \cite{ibq} & 342M & 2.88 & 254.73 & 0.84 & 0.51 \\
    & IBQ-L \cite{ibq} & 649M & 2.45 & 267.48 & 0.83 & 0.52 \\
    & IBQ-XL \cite{ibq} & 1.1B & 2.14 & 278.99 & 0.83 & 0.56 \\
    & IBQ-XXL \cite{ibq} & 2.1B & 2.05 & 286.73 & 0.83 & 0.57 \\
    & RAR-B \cite{rar} & 261M & 1.95 & 290.5 & 0.82 & 0.58 \\
    & RAR-L \cite{rar} & 461M & 1.70 & 299.5 & 0.81 & 0.60 \\
    & RAR-XL \cite{rar} & 955M & 1.50 & 306.9 & 0.80 & 0.62 \\
    & RAR-XXL \cite{rar} & 1.5B & 1.48 & 326.0 & 0.80 & 0.63 \\
    \midrule 
    \multirow{3}{*}{AR} & DAR-B (Ours) & 485M & 1.60 & 305.51 & 0.80 & 0.62 \\
    & DAR-L (Ours) & 1.1B & 1.42 & 307.94 & 0.80 & 0.63 \\
    & DAR-XL (Ours) & 2.0B & \textbf{1.37} & 320.82 & 0.81 & 0.63 \\
    \bottomrule
  \end{tabular}
  }
  \caption{\textbf{256$\times$256 ImageNet class-conditional generation results evaluated with ADM~\cite{Diff_1}.} 
  ``Type'' denotes the category of the generative model, while ``Mask.'' refers to the mask transformer model.
  ``re'' stands for rejection sampling.
  ``-384'' indicates that images are initially generated at a resolution of 384$\times$384 and subsequently resized to 256$\times$256 for evaluation.
  }
  \label{tab:compare}
\end{table}
In \cref{tab:compare}, we compare DAR with other image generation methods on class-conditional image generation task.
Our DAR significantly outperforms other autoregressive methods, \eg LlamaGen~\cite{llamagen}, Open-MAGVIT2~\cite{MAG}, IBQ~\cite{ibq} and RAR~\cite{rar}, across different model scales.
It is worth noting that DAR-B, equipped with 485M parameters and using the same image tokenizer as IBQ, achieves a remarkable FID score of 1.60.
This significantly surpasses IBQ-XXL (2.1B, FID 2.05)~\cite{ibq} while utilizing 78\% fewer parameters.
DAR-B also surpasses the autoregressive model LlamaGen-3B-384 (3.1B, FID 2.18) and the scale-wise autoregressive model FlowAR-H (1.9B, FID 1.65), while utilizing 77\% and 75\% fewer parameters, respectively.
These results demonstrate the effectiveness of our architectural optimizations for the autoregressive transformer.

As shown in \cref{tab:compare}, DAR demonstrates consistent performance improvements as the model scale increases (from 485M to 2.0B).
Our 1.1B model, DAR-L, achieves an FID score of 1.42, surpassing the performance of RAR-XXL (1.5B, FID 1.48) while utilizing 27\% fewer parameters.
Our largest proposed model, DAR-XL, achieves an FID score of 1.37, seting a new state-of-the-art result on the 256$\times$256 ImageNet benchmark~\cite{imagenet}.
Notably, RAR~\cite{rar} introduces a random-order training strategy, which is incompatible with language models and thus difficult to apply to unified multimodal generative models.
In contrast, our DAR proposes a diagonal scanning order that is utilized for both training and inference, along with enhanced autoregressive transformer modules: 4D-RoPE, direction embeddings, and codebook-based image token embeddings.
All of these components are compatible with language models, facilitating seamless integration into unified multimodal foundation models.

\Cref{tab:compare} reports the comparative results of DAR with various types of image generation methods.
DAR-XL achieves an FID score of 1.37, surpassing the variant autoregressive methods VAR (FID 1.73)~\cite{var} and MAR (FID 1.55)~\cite{mar}.
Furthermore, DAR adheres to the standard next-token prediction approach and employs a structure compatible with language models, thereby enabling it to leverage optimization and acceleration techniques designed for large language models.
Additionally, DAR outperforms the mask-based model MaskBit (FID 1.52)~\cite{mask_3} and the diffusion model MDTv2-XL/2 (FID 1.58)~\cite{cosinecfg}.
These results demonstrate that the next-token prediction paradigm, traditionally effective in language tasks, also exhibits strong capabilities in visual generation tasks.

\begin{figure*}[t]
    \centering
    \begin{subfigure}{0.3\linewidth}
        \includegraphics[width=\linewidth]{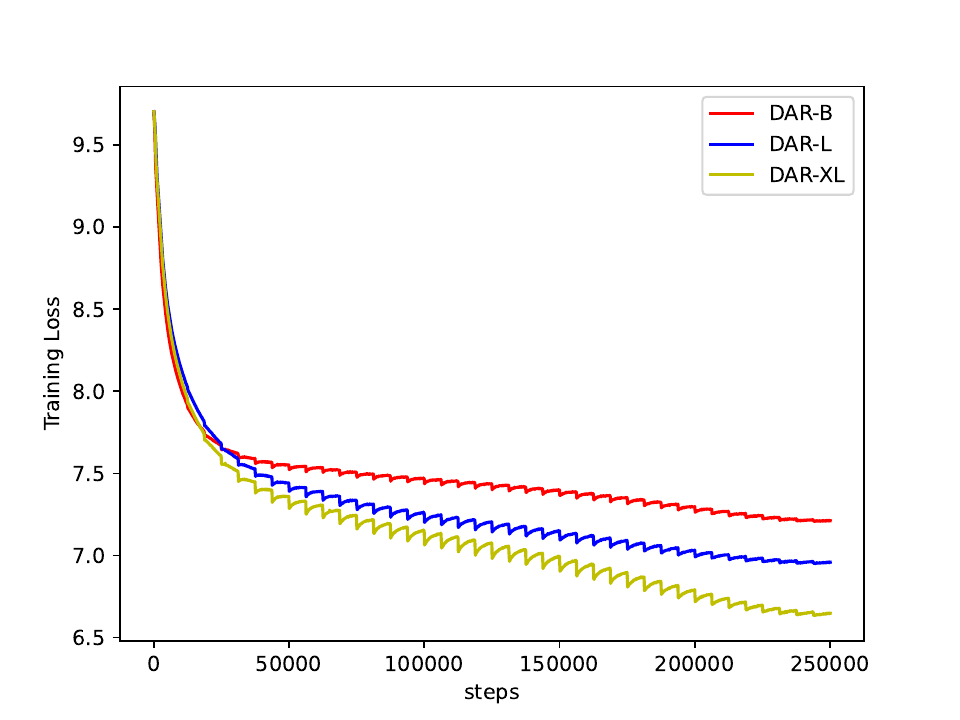}
        \caption{training losses.}
        \label{fig:loss}        
    \end{subfigure}
    \hfill
    \begin{subfigure}{0.3\linewidth}
        \includegraphics[width=\linewidth]{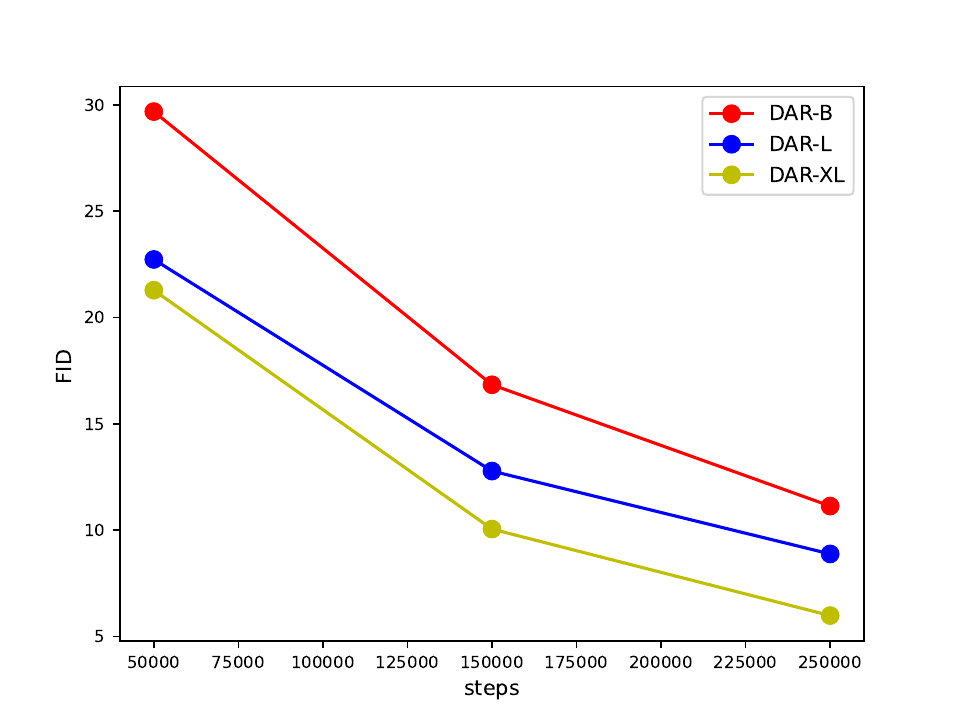}
        \caption{FID scores without classifier-free guidance}
        \label{fig:fidwocfg}        
    \end{subfigure}
    \hfill
    \begin{subfigure}{0.3\linewidth}
        \includegraphics[width=\linewidth]{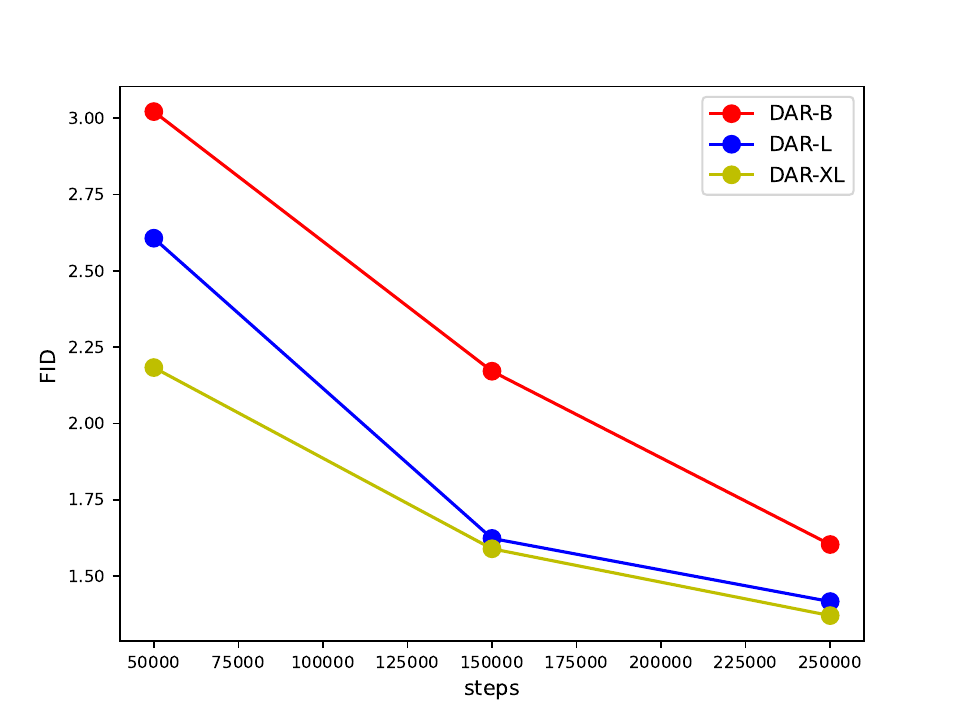}
        \caption{FID scores with classifier-freeguidance}
        \label{fig:fidcfg}        
    \end{subfigure}
    \caption{\textbf{Scaling up behavior of DAR models.} 
    We show the training loss curve for models of varying scales, alongside the FID score curves with and without classifier-free guidance.
    As the model size scales, subfigure \subref{fig:loss} demonstrates a consistent reduction in loss, while subfigures \subref{fig:fidwocfg} and \subref{fig:fidcfg} illustrate a consistent decrease in FID score without and with classifier-free guidance, respectively.
    }
    \label{fig:scaling}
\end{figure*}
\noindent
\textbf{Scaling Up Behavior.}
In \cref{fig:scaling}, we investigate the scaling effects of DAR. 
Specifically, we sample the loss every 100 steps to plot the training loss curve, and we also plot the FID score curves both with and without classifier-free guidance.
As illustrated in \cref{fig:scaling}, DAR demonstrates consistent improvements as the model scale increases, with a corresponding decrease in training loss.
Additionally, the FID scores consistently decrease, regardless of whether classifier-free guidance is used.
Notably, DAR adopts an autoregressive training paradigm, inheriting the scalability of autoregressive methods.

\begin{table}
  \centering
  \resizebox{\linewidth}{!}{
  \begin{tabular}{c|ccccc}
    \toprule
    Method & \#Para. & Precision & FID$\downarrow$ & Steps & Images/sec \\
    \midrule
    LlamaGen-XL-384~\cite{llamagen} & 775M & bf16 & 2.62 & 576 & 4.28 \\
    IBQ-XXL~\cite{ibq} & 2.1B & fp32 & 2.05 & 256 & 6.59 \\
    RAR-XXL~\cite{rar} & 1.5B & fp32 & 1.48 & 256 & 9.06 \\
    RAR-XXL*~\cite{rar} & 1.5B & bf16 & 1.48 & 256 & 13.03 \\
    DAR-L (Ours) & 1.1B & bf16 & 1.42 & 256 & 13.96 \\
    DAR-XL (Ours) & 2.0B & bf16 & 1.37 & 256 & 10.10 \\
    \bottomrule
  \end{tabular}
  }
  \caption{\textbf{Sampling throughput comparison for autoregressive transformers in different AR methods.}
  We evaluate the throughput of autoregressive transformers in different AR methods by measuring the number of samples generated per second using a batch size of 128 on a single H100 based on their official codebases.
  * specifies the precision of model parameters is converted from float32 to bfloat16.
  }
  \label{tab:speed}
\end{table}
\noindent
\textbf{Sampling Speed.}
As shown in \cref{tab:speed}, we compare the sampling speed of autoregressive transformers in different AR methods.
Our DAR-L achieves the fastest sampling speed of 13.96 images/sec while surpassing other methods in the FID score.
Specifically, DAR-L demonstrates 3.3$\times$ and 2.1$\times$ faster sampling than LlamaGen-XL-384 and IBQ-XXL respectively, while significantly outperforming them in FID metrics.
When compared with RAR-XXL under identical bfloat16 precision, DAR-L not only achieves better FID performance but also delivers faster generation speed (13.96 images/sec from DAR-L \textit{vs.} 13.03 images/sec from RAR-XXL).
DAR-XL achieves a state-of-the-art FID score of 1.37 while attaining a sampling speed of 10.10 images/sec, demonstrating 2.4$\times$ and 1.5$\times$ faster sampling than LlamaGen-XL-384 and IBQ-XXL respectively.
These results demonstrate that our enhanced autoregressive transformer modules achieve improved image quality while maintaining efficient inference capabilities.
The speed comparison of the complete sampling process including detokenization is provided in the appendix.

\begin{figure}[t]
    \centering
    \includegraphics[width=\linewidth]{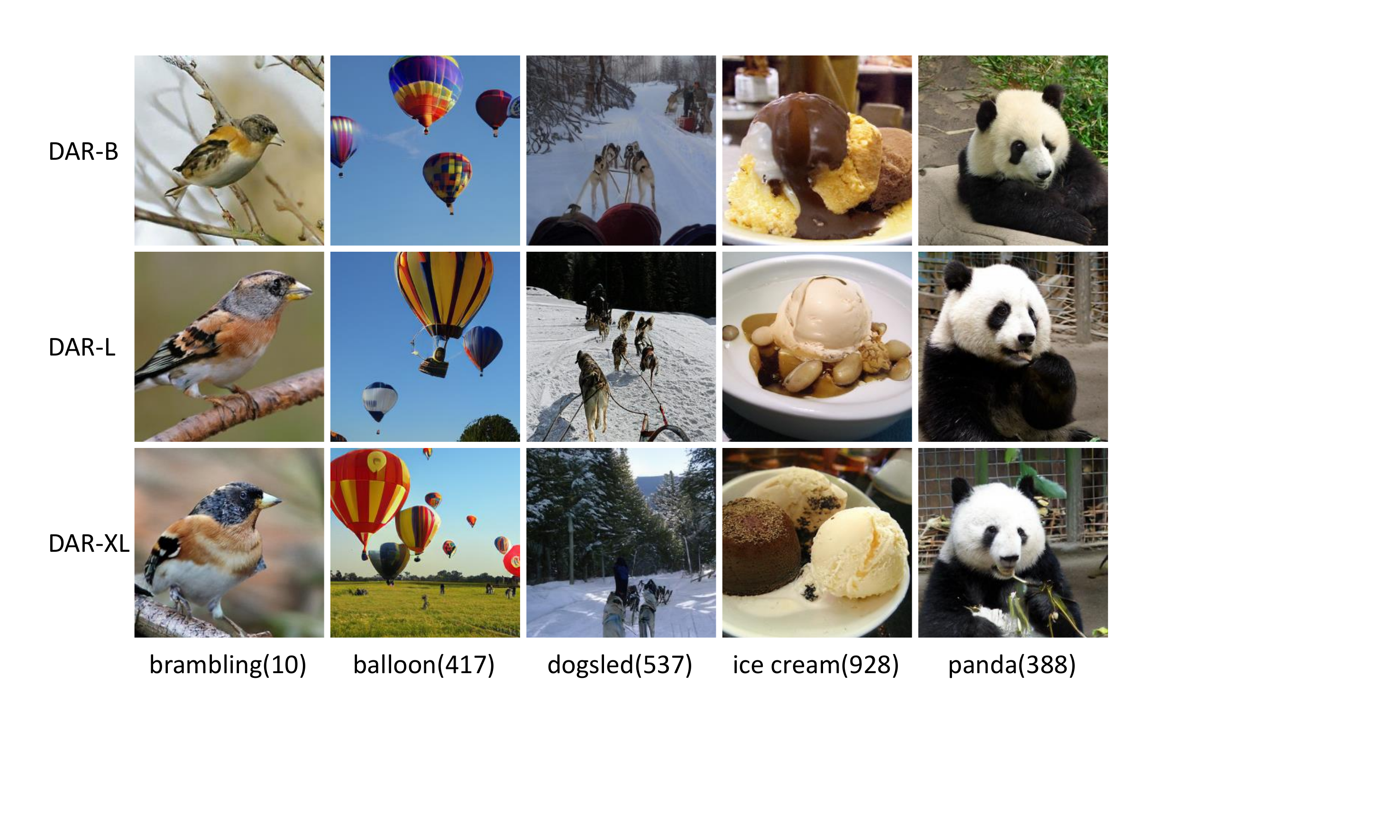}      
    \caption{\textbf{Visualization of sample images generated by DAR of varying scales.}
    As the model size increases, a noticeable improvement in image quality is observed.
    }
    \label{fig:visualization}
\end{figure}
\noindent
\textbf{Visualization.}
In \cref{fig:visualization}, we present visualizations of DAR models with varying scales, demonstrating that as the model scales, the quality of generated images improves and the diversity expands.
More visualizations can be found in appendix.


\begin{table}
  \centering
  \resizebox{\linewidth}{!}{
  \begin{tabular}{cccc|cccc}
    \toprule
    Scan Order & Code. & 4D-RoPE & Dir. & FID$\downarrow$ & IS$\uparrow$ & Pre.$\uparrow$ & Rec.$\uparrow$ \\
    \midrule
    \multirow{2}{*}{raster} & & & & 1.73 & 292.71 & 0.79 & 0.63 \\
    & $\checkmark$ & $\checkmark$ & $\checkmark$ & 1.63 & 297.80 & 0.79 & 0.62 \\
    \midrule
    \multirow{6}{*}{diagonal} & & & & 1.78 & 303.17 & 0.79 & 0.61 \\
    & & $\checkmark$ & $\checkmark$ & 1.71 & 300.77 & 0.80 & 0.62 \\
    & $\checkmark$ & & & 1.76 & 301.44 & 0.79 & 0.63 \\
    & $\checkmark$ & $\checkmark$ & & 1.62 & 304.71 & 0.80 & 0.62 \\
    & $\checkmark$ & & $\checkmark$ & 1.63 & 296.53 & 0.79 & 0.62 \\
    & $\checkmark$ & $\checkmark$ & $\checkmark$ & \textbf{1.60} & 305.51 & 0.80 & 0.62 \\
    \bottomrule
  \end{tabular}
  }
  \caption{\textbf{Effectiveness of designed modules.}
  ``Code.'' refers to codebook-based image token embeddings.
  ``Dir.'' denotes direction embeddings.
  We conduct experiments on the model size DAR-B using two distinct image token scanning orders and varying model configurations.
  All models are trained with ten-crop data augmentation and use classifier-free guidance~\cite{cfg} combined with the power-cosine guidance schedule~\cite{cosinecfg} for sampling.}
  \label{tab:ablation}
\end{table}
\subsection{Ablation Studies}
\label{subsec:ablation}
To validate the effectiveness of our method, we conduct ablation studies on the image token scanning order and several key modules, as shown in \cref{tab:ablation}.

\noindent
\textbf{Effect of diagonal scanning order.}
In \cref{tab:ablation}, we compare the results using raster scan order and diagonal scan order.
When employing direction-aware modules (4D-RoPE and direction embeddings) and the image tokenizer codebook, the diagonal scan strategy achieves a superior FID score of 1.60 compared to the raster scan order's 1.63.
The results demonstrate that the diagonal scanning order, which ensures all adjacent tokens are in close proximity and provides a broader field of view, offers significant advantages.
Notably, when the proposed modules are not utilized, the raster scan order achieves an FID of 1.73, demonstrating a significant advantage over the diagonal scan order, which yields an FID of 1.78. 
This discrepancy arises because the diagonal scan order has more generation directions and more frequent changes in generation direction compared to the raster scan order, which is challenging for vanilla autoregressive transformers.
This observation further underscores the necessity of the direction-aware modules.

\noindent
\textbf{Effect of direction-aware modules.}
In \cref{tab:ablation}, we validate the effectiveness of our proposed two direction-aware modules: 4D-RoPE and direction embeddings.
When employing the diagonal scanning order and the codebook-based image token embeddings, the model without any direction-aware module achieves an FID of 1.76.
By solely incorporating 4D-RoPE, the FID improves to 1.62, while using only direction embeddings reduces the FID to 1.63.
These results demonstrate that both 4D-RoPE and direction embeddings individually improve the model's capability to handle different generation directions, with 4D-RoPE showing a more significant improvement (FID 1.62 vs. FID 1.63).
Specifically, 4D-RoPE injects directional information into the attention matrix, whereas direction embeddings integrate such information into AdaLN.
When both modules are utilized simultaneously, the FID is further improved to 1.60, indicating that the two modules are compatible with each other.
Furthermore, when employing the raster scan order, the baseline model achieves an FID of 1.73.
With the integration of all proposed modules, the FID is reduced to 1.63, demonstrating the adaptability of our modules to different scanning orders.
\begin{table}
  \centering
  \resizebox{\linewidth}{!}{
  \begin{tabular}{c|cccc}
    \toprule
    Adaln Condition & FID$\downarrow$ & IS$\uparrow$ & Pre.$\uparrow$ & Rec.$\uparrow$ \\
    \midrule
    class & 1.618 & 304.71 & 0.80 & 0.62 \\
    class + timestep & 1.611 & 293.67 & 0.79 & 0.63 \\
    class + direction & \textbf{1.603} & 305.51 & 0.80 & 0.62 \\
    \bottomrule
  \end{tabular}
  }
  \caption{\textbf{Different conditions for AdaLN.}
  ``class'' refers to the class embedding.
  ``class + timestep'' refers to the sum of the class embedding and the generation step embedding.
  ``class + direction'' refers to the sum of the class embedding and the direction embedding.
  We compare the results of using different conditions within AdaLN.
  }
  \label{tab:adaln}
\end{table}

In \cref{tab:adaln}, we present a comparison of different Adaln conditions.
Utilizing only the class embedding achieves an FID score of 1.618.
When combining class embedding and timestep embedding, the FID score marginally improves to 1.611, indicating a limited enhancement in performance.
This is because, unlike in diffusion steps, the model inherently distinguishes generation steps through causal attention, rendering timestep information redundant.
Conversely, the sum of the class embedding and the direction embedding achieves an FID score of 1.603.
This improvement suggests that the model initially lacks direction awareness, which is compensated by the direction embedding.

\noindent
\textbf{Effect of codebook-based image token embeddings.}
In \cref{tab:ablation}, we validate the effectiveness of codebook-based image token embeddings.
When employing the diagonal scanning order and without any direction-aware modules, the model utilizing the codebook for image token embeddings achieves an FID score of 1.76, slightly outperforming the model learning image token embeddings from scratch, which achieves an FID score of 1.78.
This indicates that conventional autoregressive transformers, when learning generative tasks, exhibit limited dependence on pretrained embeddings.
However, when direction-aware modules are incorporated, the model utilizing the codebook achieves an FID score of 1.60, significantly surpassing the FID score of 1.71 achieved by the model learning embeddings from scratch.
This demonstrates that well-trained image token representations greatly enhance the model's ability to perceive directional information in generative tasks.
These results underscore the necessity of codebook-based image token embeddings in direction-aware transformers.

%% file: sec/5_conclusion.tex
\section{Conclusion}
\label{sec:conclusion}
In this paper, we introduce the diagonal scanning order for next-token prediction and propose a direction-aware autoregressive transformer framework.
Our method ensures that tokens with adjacent indices are positioned in close proximity.
Furthermore, direction-aware modules enhance the model's capability to handle frequent changes in generation direction.
We propose models at multiple scales ranging from 485M to 2.0B.
Our DAR achieves an FID score of 1.37, surpassing previous state-of-the-art autoregressive approaches.
We hope this research contributes to the field of unified multimodal foundation models.

%% file: sec/X_suppl.tex
\maketitlesupplementary

\section*{Appendix}
\label{sec:appendix}

\section*{A. Hyper-parameters for DAR Models}
\label{sec:hyper}
We list the detailed training and sampling hyper-parameters for DAR models in \cref{tab:hyper}.
\begin{table}[h!]
\centering
  \resizebox{\linewidth}{!}{
  \begin{tabular}{l|c}
    \toprule
    config \quad\quad\quad\quad\quad\quad\quad\quad & value \\
    \midrule
    \multicolumn{2}{c}{\textit{training hyper-parameters}} \\
    \midrule
    optimizer & AdamW~\cite{adamw_1, adamw_2} \\
    learning rate & 1e-3 (B) / 4e-4 (L, XL) \\
    weight decay & 0.05 \\
    optimizer momentum & (0.9, 0.96) \\
    batch size & 2048 \\
    learning rate schedule & cosine decay \\
    ending learning rate & 1e-5 \\
    total epochs & 400 \\
    warmup epochs & 100 \\
    precision & bfloat16 \\
    max grad norm & 1.0 \\
    dropout rate & 0.1 \\
    attn dropout rate & 0.1 \\
    class label dropout rate & 0.1 \\
    \midrule
    \multicolumn{2}{c}{\textit{sampling hyper-parameters}}  \\
    \midrule
    guidance schedule & pow-cosine~\cite{cosinecfg} \\
    temperature & 1.02 (B) / 1.04 (L) / 1.02 (XL) \\
    scale power & 0.88 (B) / 0.78 (L) / 0.56 (XL) \\
    guidance scale & 4.7 (B) / 4.5 (L) / 4.3 (XL) \\
    \bottomrule
    \end{tabular}
    }
    \caption{\textbf{Detailed hyper-parameters for DAR models.}}
    \label{tab:hyper}
\end{table}

\section*{B. Speed Comparison of the Complete Sampling Process}
\label{sec:speed_comp}
\begin{table}[h!]
  \centering
  \resizebox{\linewidth}{!}{
  \begin{tabular}{c|ccccc}
    \toprule
    Method & \#Para. & Precision & FID$\downarrow$ & Steps & Images/sec \\
    \midrule
    LlamaGen-XL-384~\cite{llamagen} & 1.4B & bf16 & 2.34 & 384 & 4.20 \\
    IBQ-XXL~\cite{ibq} & 2.1B & fp32 & 2.05 & 256 & 6.32 \\
    RAR-XXL~\cite{rar} & 1.5B & fp32 & 1.48 & 256 & 8.86 \\
    RAR-XXL*~\cite{rar} & 1.5B & bf16 & 1.48 & 256 & 12.63 \\
    DAR-L (Ours) & 1.1B & bf16 & 1.42 & 256 & 13.45 \\
    DAR-XL (Ours) & 2.0B & bf16 & 1.37 & 256 & 9.80 \\
    \bottomrule
  \end{tabular}
  }
  \caption{\textbf{Sampling throughput comparison for the complete sampling process.}
  We evaluate the throughput of the complete sampling process by measuring the number of samples generated per second using a batch size of 128 on a single H100 based on their official codebases.
  * specifies the precision of model parameters is converted from float32 to bfloat16.
  }
  \label{tab:speedfull}
\end{table}

As shown in \cref{tab:speedfull}, our DAR-L achieves the fastest sampling speed of 13.45 images/sec while surpassing other methods in the FID score.
Specifically, DAR-L achieves a sampling speed of 13.45 images/sec, which is 3.2$\times$ faster than LlamaGen-XL-384~\cite{llamagen} and 2.1$\times$ faster than IBQ-XXL~\cite{ibq}.
Moreover, DAR-L significantly outperforms both LlamaGen-XL-384 and IBQ-XXL in terms of FID score.
When compared with RAR-XXL~\cite{rar} under the same bfloat16 precision, DAR-L demonstrates superior sampling speed (13.45 images/sec from DAR-L \textit{vs.} 12.63 images/sec from RAR-XXL) and better image generation quality (FID 1.42 from DAR-L \textit{vs.} FID 1.48 from RAR-XXL).
Our proposed DAR-XL achieves the state-of-the-art FID score of 1.37, with a sampling speed of 9.80 images/sec, which is 2.3$\times$ and 1.6$\times$ faster than LlamaGen-XL-384 and IBQ-XXL, respectively.
These results demonstrate that the proposed 4D-RoPE module, direction embeddings and the codebook-based image token embeddings, when applied to the autoregressive transformer for handling frequently changing generation directions, not only significantly enhance the quality of generated images but also maintain the model's high-speed inference capabilities.

\section*{C. Additional Visualizations}
\label{sec:more_visual}
\begin{figure}
    \centering
    \includegraphics[width=\linewidth]{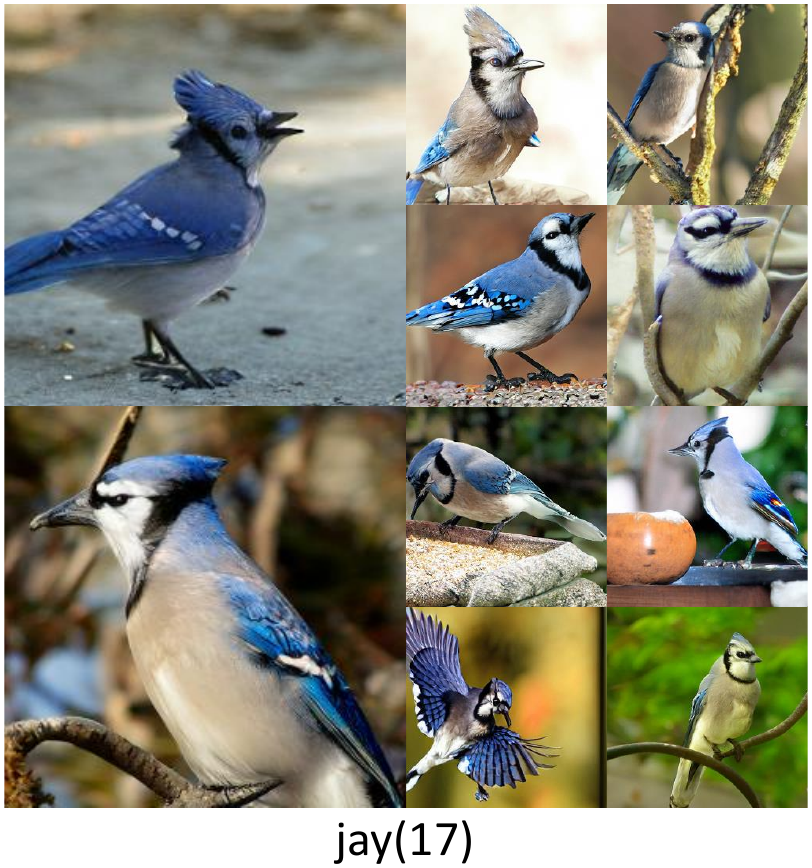}      
    \caption{\textbf{Generated samples from DAR.}
    DAR is capable of generating high-fidelity and highly diverse images.}
    \label{fig:jay17}
\end{figure}

\begin{figure}
    \centering
    \includegraphics[width=\linewidth]{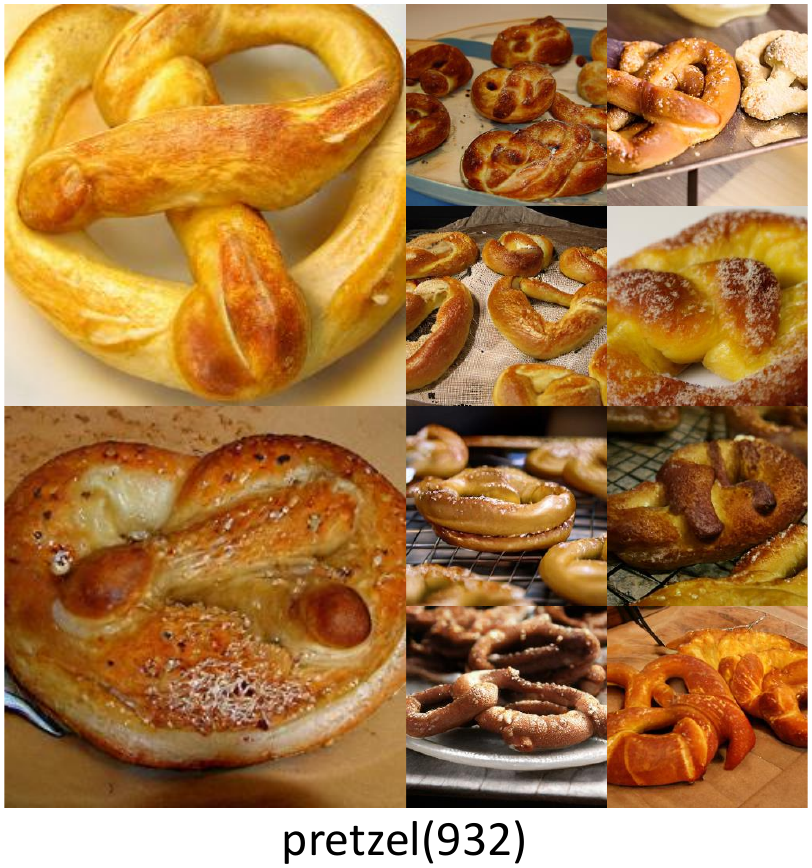}      
    \caption{\textbf{Generated samples from DAR.}
    DAR is capable of generating high-fidelity and highly diverse images.}
    \label{fig:pretzel932}
\end{figure}

\begin{figure}
    \centering
    \includegraphics[width=\linewidth]{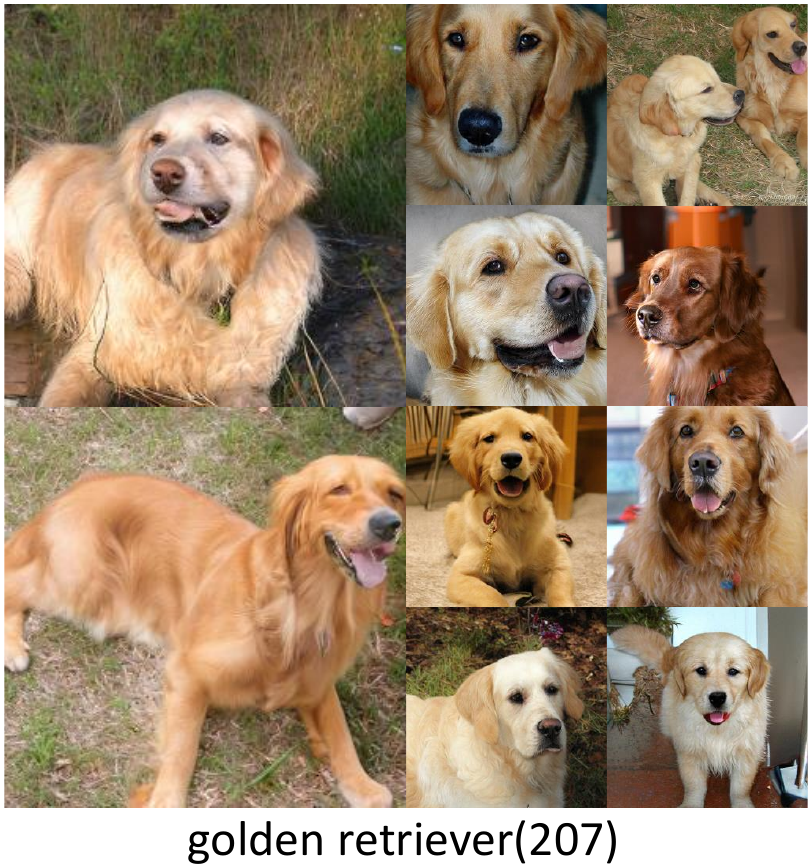}      
    \caption{\textbf{Generated samples from DAR.}
    DAR is capable of generating high-fidelity and highly diverse images.}
    \label{fig:golden_retriever207}
\end{figure}

\begin{figure}
    \centering
    \includegraphics[width=\linewidth]{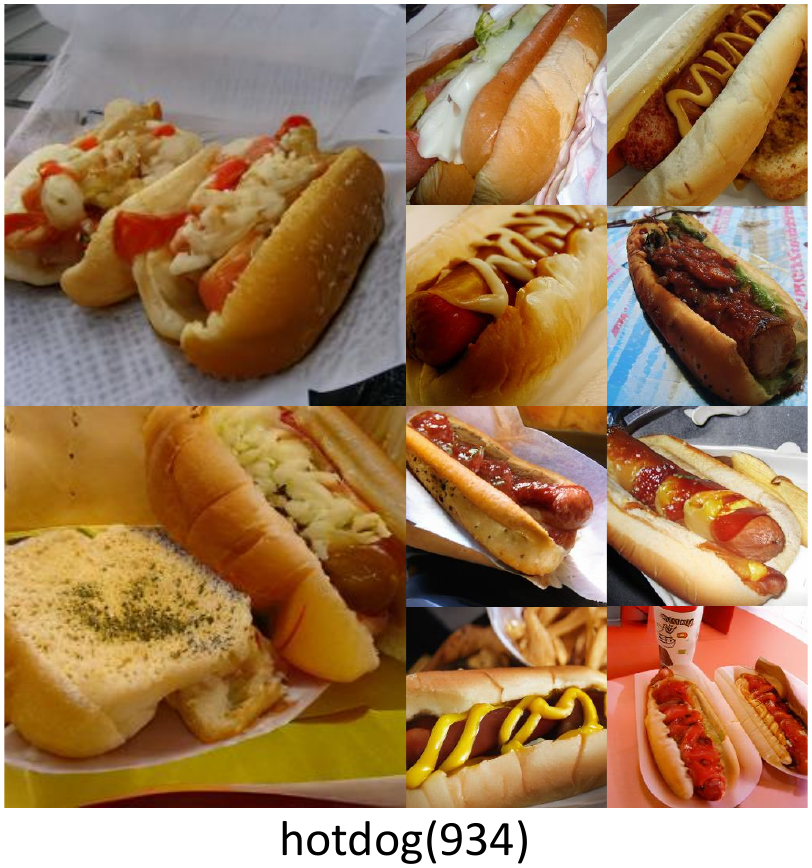}      
    \caption{\textbf{Generated samples from DAR.}
    DAR is capable of generating high-fidelity and highly diverse images.}
    \label{fig:hotdog934}
\end{figure}

\begin{figure}
    \centering
    \includegraphics[width=\linewidth]{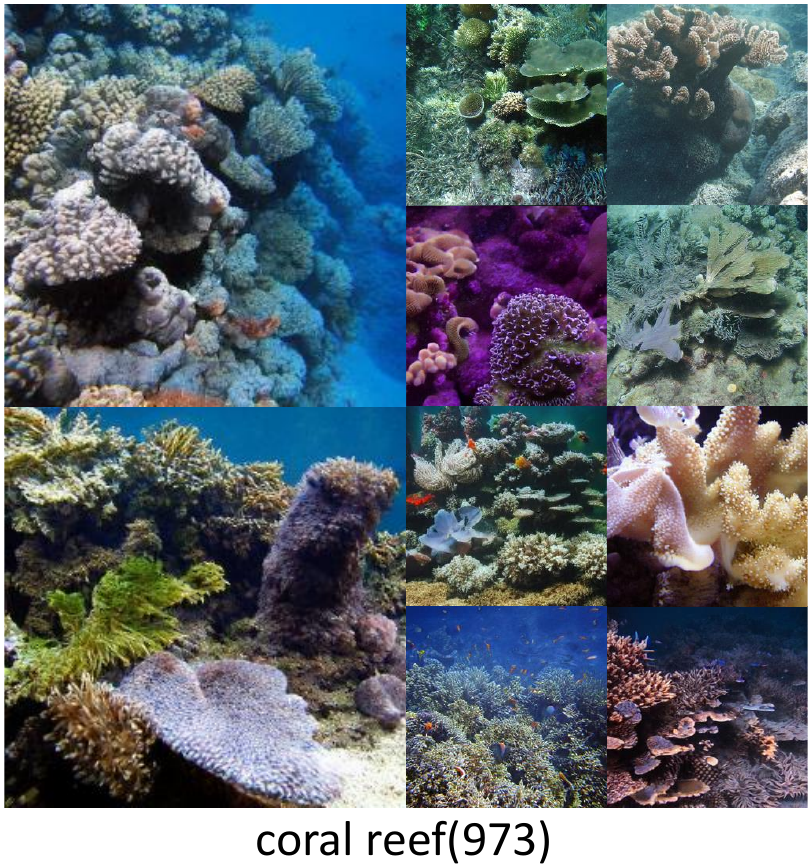}      
    \caption{\textbf{Generated samples from DAR.}
    DAR is capable of generating high-fidelity and highly diverse images.}
    \label{fig:coral973}
\end{figure}

\begin{figure}
    \centering
    \includegraphics[width=\linewidth]{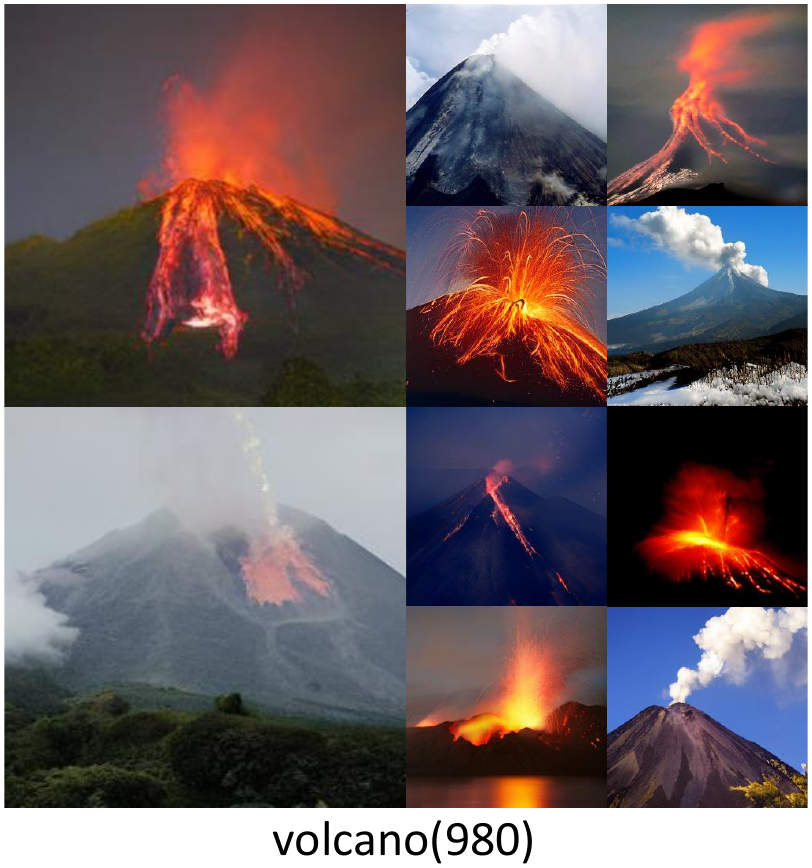}      
    \caption{\textbf{Generated samples from DAR.}
    DAR is capable of generating high-fidelity and highly diverse images.}
    \label{fig:volcano980}
\end{figure}

\begin{figure}
    \centering
    \includegraphics[width=\linewidth]{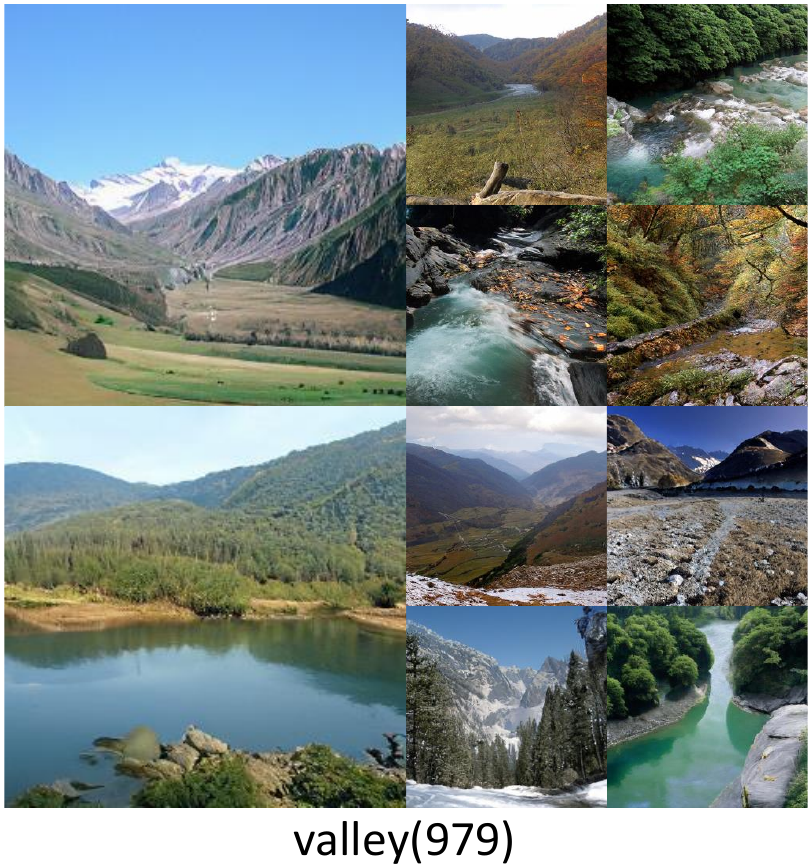}      
    \caption{\textbf{Generated samples from DAR.}
    DAR is capable of generating high-fidelity and highly diverse images.}
    \label{fig:valley979}
\end{figure}

\begin{figure}
    \centering
    \includegraphics[width=\linewidth]{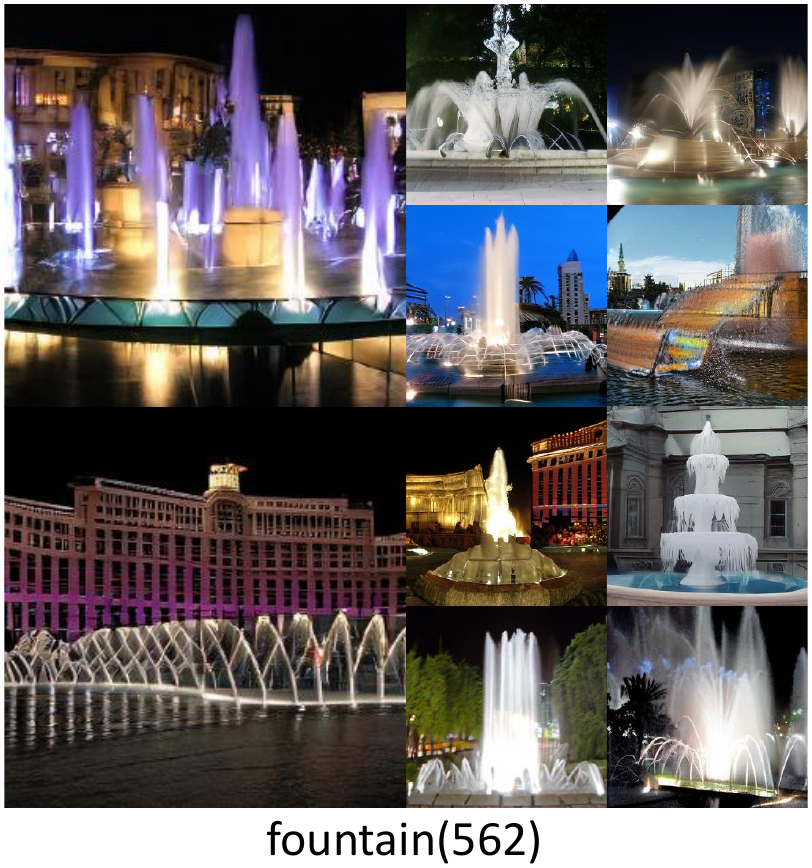}      
    \caption{\textbf{Generated samples from DAR.}
    DAR is capable of generating high-fidelity and highly diverse images.}
    \label{fig:fountain562}
\end{figure}

We provide additional visualization results generated by DAR in \cref{fig:jay17}, \cref{fig:pretzel932}, \cref{fig:golden_retriever207}, \cref{fig:hotdog934}, \cref{fig:coral973}, \cref{fig:volcano980}, \cref{fig:valley979} and \cref{fig:fountain562}.

\section*{D. Limitations and Future Work}
\label{sec:lim}
In this work, we primarily focus on architectural improvements to the autoregressive transformer.
Our approach maintains compatibility with any discrete tokenizer.
We believe that incorporating more advanced image tokenizers can further enhance the performance of autoregressive visual generation.
We train our model on ImageNet~\cite{imagenet}, specifically focusing on class-conditional generation.
The generative performance could be further enhanced by leveraging image-text paired data.
Our enhanced autoregressive transformer is also compatible with unified multimodal generative models.
We leave these for future work.

%% file: main.bbl
\begin{thebibliography}{68}
\providecommand{\natexlab}[1]{#1}
\providecommand{\url}[1]{\texttt{#1}}
\expandafter\ifx\csname urlstyle\endcsname\relax
  \providecommand{\doi}[1]{doi: #1}\else
  \providecommand{\doi}{doi: \begingroup \urlstyle{rm}\Url}\fi

\bibitem[Achiam et~al.(2023)Achiam, Adler, Agarwal, Ahmad, Akkaya, Aleman, Almeida, Altenschmidt, Altman, Anadkat, et~al.]{llm_3}
Josh Achiam, Steven Adler, Sandhini Agarwal, Lama Ahmad, Ilge Akkaya, Florencia~Leoni Aleman, Diogo Almeida, Janko Altenschmidt, Sam Altman, Shyamal Anadkat, et~al.
\newblock Gpt-4 technical report.
\newblock \emph{arXiv preprint arXiv:2303.08774}, 2023.

\bibitem[Bai et~al.(2024)Bai, Geng, Mangalam, Bar, Yuille, Darrell, Malik, and Efros]{vlm_7}
Yutong Bai, Xinyang Geng, Karttikeya Mangalam, Amir Bar, Alan Yuille, Trevor Darrell, Jitendra Malik, and Alexei~A Efros.
\newblock Sequential modeling enables scalable learning for large vision models.
\newblock In \emph{CVPR}, 2024.

\bibitem[Brock et~al.(2018)Brock, Donahue, and Simonyan]{GAN_2}
Andrew Brock, Jeff Donahue, and Karen Simonyan.
\newblock Large scale {GAN} training for high fidelity natural image synthesis.
\newblock \emph{CoRR}, abs/1809.11096, 2018.

\bibitem[Chang et~al.(2022)Chang, Zhang, Jiang, Liu, and Freeman]{mask_1}
Huiwen Chang, Han Zhang, Lu Jiang, Ce Liu, and William~T. Freeman.
\newblock Maskgit: Masked generative image transformer.
\newblock In \emph{Proceedings of the IEEE/CVF Conference on Computer Vision and Pattern Recognition (CVPR)}, pages 11315--11325, 2022.

\bibitem[Chang et~al.(2023)Chang, Zhang, Barber, Maschinot, Lezama, Jiang, Yang, Murphy, Freeman, Rubinstein, Li, and Krishnan]{mask_2}
Huiwen Chang, Han Zhang, Jarred Barber, AJ Maschinot, Jos\'{e} Lezama, Lu Jiang, Ming-Hsuan Yang, Kevin Murphy, William~T. Freeman, Michael Rubinstein, Yuanzhen Li, and Dilip Krishnan.
\newblock Muse: Text-to-image generation via masked generative transformers.
\newblock In \emph{Proceedings of the 40th International Conference on Machine Learning}. JMLR.org, 2023.

\bibitem[Chen et~al.(2020)Chen, Radford, Child, Wu, Jun, Luan, and Sutskever]{ar_1}
Mark Chen, Alec Radford, Rewon Child, Jeff Wu, Heewoo Jun, David Luan, and Ilya Sutskever.
\newblock Generative pretraining from pixels.
\newblock In \emph{Proceedings of the 37th International Conference on Machine Learning}. JMLR.org, 2020.

\bibitem[Dai et~al.(2023)Dai, Li, Li, Tiong, Zhao, Wang, Li, Fung, and Hoi]{vlm_3}
Wenliang Dai, Junnan Li, Dongxu Li, Anthony Tiong, Junqi Zhao, Weisheng Wang, Boyang Li, Pascale Fung, and Steven Hoi.
\newblock Instruct{BLIP}: Towards general-purpose vision-language models with instruction tuning.
\newblock In \emph{Thirty-seventh Conference on Neural Information Processing Systems}, 2023.

\bibitem[Deng et~al.(2009)Deng, Dong, Socher, Li, Li, and Fei-Fei]{imagenet}
Jia Deng, Wei Dong, Richard Socher, Li-Jia Li, Kai Li, and Li Fei-Fei.
\newblock Imagenet: A large-scale hierarchical image database.
\newblock In \emph{2009 IEEE Conference on Computer Vision and Pattern Recognition}, pages 248--255, 2009.

\bibitem[Devlin et~al.(2018)Devlin, Chang, Lee, and Toutanova]{ape_1}
Jacob Devlin, Ming-Wei Chang, Kenton Lee, and Kristina Toutanova.
\newblock Bert: Pre-training of deep bidirectional transformers for language understanding.
\newblock In \emph{NAACL}, 2018.

\bibitem[Dhariwal and Nichol(2021)]{Diff_1}
Prafulla Dhariwal and Alexander Nichol.
\newblock Diffusion models beat gans on image synthesis.
\newblock In \emph{Advances in Neural Information Processing Systems}, pages 8780--8794. Curran Associates, Inc., 2021.

\bibitem[Dosovitskiy et~al.(2021)Dosovitskiy, Beyer, Kolesnikov, Weissenborn, Zhai, Unterthiner, Dehghani, Minderer, Heigold, Gelly, Uszkoreit, and Houlsby]{ape_2}
Alexey Dosovitskiy, Lucas Beyer, Alexander Kolesnikov, Dirk Weissenborn, Xiaohua Zhai, Thomas Unterthiner, Mostafa Dehghani, Matthias Minderer, Georg Heigold, Sylvain Gelly, Jakob Uszkoreit, and Neil Houlsby.
\newblock An image is worth 16x16 words: Transformers for image recognition at scale.
\newblock In \emph{ICLR}, 2021.

\bibitem[Esser et~al.(2021)Esser, Rombach, and Ommer]{ar_4}
Patrick Esser, Robin Rombach, and Bjorn Ommer.
\newblock Taming transformers for high-resolution image synthesis.
\newblock In \emph{Proceedings of the IEEE/CVF Conference on Computer Vision and Pattern Recognition (CVPR)}, pages 12873--12883, 2021.

\bibitem[Esser et~al.(2024)Esser, Kulal, Blattmann, Entezari, M\"{u}ller, Saini, Levi, Lorenz, Sauer, Boesel, Podell, Dockhorn, English, and Rombach]{Diff_2}
Patrick Esser, Sumith Kulal, Andreas Blattmann, Rahim Entezari, Jonas M\"{u}ller, Harry Saini, Yam Levi, Dominik Lorenz, Axel Sauer, Frederic Boesel, Dustin Podell, Tim Dockhorn, Zion English, and Robin Rombach.
\newblock Scaling rectified flow transformers for high-resolution image synthesis.
\newblock In \emph{Proceedings of the 41st International Conference on Machine Learning}. JMLR.org, 2024.

\bibitem[Fang et~al.(2024)Fang, Sun, Wang, Huang, Wang, and Cao]{eva02}
Yuxin Fang, Quan Sun, Xinggang Wang, Tiejun Huang, Xinlong Wang, and Yue Cao.
\newblock Eva-02: A visual representation for neon genesis.
\newblock \emph{Image and Vision Computing}, 149:\penalty0 105171, 2024.

\bibitem[Gao et~al.(2023)Gao, Zhou, Cheng, and Yan]{cosinecfg}
Shanghua Gao, Pan Zhou, Ming-Ming Cheng, and Shuicheng Yan.
\newblock Masked diffusion transformer is a strong image synthesizer.
\newblock In \emph{Proceedings of the IEEE/CVF International Conference on Computer Vision (ICCV)}, pages 23164--23173, 2023.

\bibitem[Goodfellow et~al.(2014)Goodfellow, Pouget{-}Abadie, Mirza, Xu, Warde{-}Farley, Ozair, Courville, and Bengio]{GAN}
Ian~J. Goodfellow, Jean Pouget{-}Abadie, Mehdi Mirza, Bing Xu, David Warde{-}Farley, Sherjil Ozair, Aaron~C. Courville, and Yoshua Bengio.
\newblock Generative adversarial nets.
\newblock In \emph{Advances in Neural Information Processing Systems 27: Annual Conference on Neural Information Processing Systems 2014, December 8-13 2014, Montreal, Quebec, Canada}, pages 2672--2680, 2014.

\bibitem[Gregor et~al.(2014)Gregor, Danihelka, Mnih, Blundell, and Wierstra]{ar_2}
Karol Gregor, Ivo Danihelka, Andriy Mnih, Charles Blundell, and Daan Wierstra.
\newblock Deep autoregressive networks.
\newblock In \emph{International Conference on Machine Learning}, pages 1242--1250. PMLR, 2014.

\bibitem[Heo et~al.(2024)Heo, Park, Han, and Yun]{mixrope}
Byeongho Heo, Song Park, Dongyoon Han, and Sangdoo Yun.
\newblock Rotary position embedding for vision transformer.
\newblock In \emph{Computer Vision – ECCV 2024: 18th European Conference, Milan, Italy, September 29–October 4, 2024, Proceedings, Part X}, page 289–305, Berlin, Heidelberg, 2024. Springer-Verlag.

\bibitem[Heusel et~al.(2017)Heusel, Ramsauer, Unterthiner, Nessler, and Hochreiter]{rfid}
Martin Heusel, Hubert Ramsauer, Thomas Unterthiner, Bernhard Nessler, and Sepp Hochreiter.
\newblock Gans trained by a two time-scale update rule converge to a local nash equilibrium.
\newblock In \emph{Proceedings of the 31st International Conference on Neural Information Processing Systems}, page 6629–6640, Red Hook, NY, USA, 2017. Curran Associates Inc.

\bibitem[Ho and Salimans(2022)]{cfg}
Jonathan Ho and Tim Salimans.
\newblock Classifier-free diffusion guidance.
\newblock \emph{arXiv preprint arXiv:2207.12598}, 2022.

\bibitem[Ho et~al.(2020)Ho, Jain, and Abbeel]{Diff_3}
Jonathan Ho, Ajay Jain, and Pieter Abbeel.
\newblock Denoising diffusion probabilistic models.
\newblock In \emph{Advances in Neural Information Processing Systems}, pages 6840--6851. Curran Associates, Inc., 2020.

\bibitem[Karras et~al.(2019)Karras, Laine, and Aila]{Karras_2019_CVPR}
Tero Karras, Samuli Laine, and Timo Aila.
\newblock A style-based generator architecture for generative adversarial networks.
\newblock In \emph{Proceedings of the IEEE/CVF Conference on Computer Vision and Pattern Recognition (CVPR)}, 2019.

\bibitem[Kingma and Ba(2015)]{adamw_1}
Diederik~P Kingma and Jimmy Ba.
\newblock Adam: A method for stochastic optimization.
\newblock In \emph{ICLR}, 2015.

\bibitem[Kynk{\"{a}}{\"{a}}nniemi et~al.(2019)Kynk{\"{a}}{\"{a}}nniemi, Karras, Laine, Lehtinen, and Aila]{prerec}
Tuomas Kynk{\"{a}}{\"{a}}nniemi, Tero Karras, Samuli Laine, Jaakko Lehtinen, and Timo Aila.
\newblock Improved precision and recall metric for assessing generative models.
\newblock In \emph{Advances in Neural Information Processing Systems 32: Annual Conference on Neural Information Processing Systems 2019, NeurIPS 2019, December 8-14, 2019, Vancouver, BC, Canada}, pages 3929--3938, 2019.

\bibitem[Lee et~al.(2022)Lee, Kim, Kim, Cho, and Han]{ar_9}
Doyup Lee, Chiheon Kim, Saehoon Kim, Minsu Cho, and Wook-Shin Han.
\newblock Autoregressive image generation using residual quantization.
\newblock In \emph{Proceedings of the IEEE/CVF Conference on Computer Vision and Pattern Recognition (CVPR)}, pages 11523--11532, 2022.

\bibitem[Li et~al.(2024)Li, Tian, Li, Deng, and He]{mar}
Tianhong Li, Yonglong Tian, He Li, Mingyang Deng, and Kaiming He.
\newblock Autoregressive image generation without vector quantization.
\newblock In \emph{Advances in Neural Information Processing Systems}, pages 56424--56445. Curran Associates, Inc., 2024.

\bibitem[Liu et~al.(2023)Liu, Li, Wu, and Lee]{vlm_1}
Haotian Liu, Chunyuan Li, Qingyang Wu, and Yong~Jae Lee.
\newblock Visual instruction tuning.
\newblock \emph{NeurIPS}, 2023.

\bibitem[Liu et~al.(2024)Liu, Zeng, He, Yu, Shen, and Chen]{Diff_4}
Qihao Liu, Zhanpeng Zeng, Ju He, Qihang Yu, Xiaohui Shen, and Liang-Chieh Chen.
\newblock Alleviating distortion in image generation via multi-resolution diffusion models and time-dependent layer normalization.
\newblock In \emph{The Thirty-eighth Annual Conference on Neural Information Processing Systems}, 2024.

\bibitem[Liu et~al.(2021)Liu, Lin, Cao, Hu, Wei, Zhang, Lin, and Guo]{rpb_1}
Ze Liu, Yutong Lin, Yue Cao, Han Hu, Yixuan Wei, Zheng Zhang, Stephen Lin, and Baining Guo.
\newblock Swin transformer: Hierarchical vision transformer using shifted windows.
\newblock In \emph{ICCV}, 2021.

\bibitem[Loshchilov and Hutter(2019)]{adamw_2}
Ilya Loshchilov and Frank Hutter.
\newblock Decoupled weight decay regularization.
\newblock \emph{ICLR}, 2019.

\bibitem[Lu et~al.(2023)Lu, Clark, Zellers, Mottaghi, and Kembhavi]{vlm_6}
Jiasen Lu, Christopher Clark, Rowan Zellers, Roozbeh Mottaghi, and Aniruddha Kembhavi.
\newblock {UNIFIED-IO:} {A} unified model for vision, language, and multi-modal tasks.
\newblock In \emph{The Eleventh International Conference on Learning Representations, {ICLR} 2023, Kigali, Rwanda, May 1-5, 2023}. OpenReview.net, 2023.

\bibitem[Lu et~al.(2024{\natexlab{a}})Lu, Clark, Lee, Zhang, Khosla, Marten, Hoiem, and Kembhavi]{2drope_1}
Jiasen Lu, Christopher Clark, Sangho Lee, Zichen Zhang, Savya Khosla, Ryan Marten, Derek Hoiem, and Aniruddha Kembhavi.
\newblock Unified-io 2: Scaling autoregressive multimodal models with vision, language, audio, and action.
\newblock In \emph{{IEEE/CVF} Conference on Computer Vision and Pattern Recognition, {CVPR} 2024, Seattle, WA, USA, June 16-22, 2024}, pages 26429--26445. {IEEE}, 2024{\natexlab{a}}.

\bibitem[Lu et~al.(2024{\natexlab{b}})Lu, Wang, Huang, Wu, Liu, Ouyang, and Bai]{2drope_2}
Zeyu Lu, Zidong Wang, Di Huang, Chengyue Wu, Xihui Liu, Wanli Ouyang, and Lei Bai.
\newblock Fit: Flexible vision transformer for diffusion model.
\newblock In \emph{Forty-first International Conference on Machine Learning, {ICML} 2024, Vienna, Austria, July 21-27, 2024}. OpenReview.net, 2024{\natexlab{b}}.

\bibitem[Luo et~al.(2024)Luo, Shi, Ge, Yang, Wang, and Shan]{MAG}
Zhuoyan Luo, Fengyuan Shi, Yixiao Ge, Yujiu Yang, Limin Wang, and Ying Shan.
\newblock Open-magvit2: An open-source project toward democratizing auto-regressive visual generation.
\newblock \emph{arXiv preprint arXiv:2409.04410}, 2024.

\bibitem[Ma et~al.(2024)Ma, Jiang, Wu, Yuan, and Qi]{vlm_4}
Chuofan Ma, Yi Jiang, Jiannan Wu, Zehuan Yuan, and Xiaojuan Qi.
\newblock Groma: Localized visual tokenization for grounding multimodal large language models.
\newblock In \emph{Computer Vision - {ECCV} 2024 - 18th European Conference, Milan, Italy, September 29-October 4, 2024, Proceedings, Part {VI}}, pages 417--435. Springer, 2024.

\bibitem[Parmar et~al.(2018)Parmar, Vaswani, Uszkoreit, Kaiser, Shazeer, Ku, and Tran]{ar_3}
Niki Parmar, Ashish Vaswani, Jakob Uszkoreit, Lukasz Kaiser, Noam Shazeer, Alexander Ku, and Dustin Tran.
\newblock Image transformer.
\newblock In \emph{Proceedings of the 35th International Conference on Machine Learning}, pages 4055--4064. PMLR, 2018.

\bibitem[Peebles and Xie(2023)]{adaln}
William Peebles and Saining Xie.
\newblock Scalable diffusion models with transformers.
\newblock In \emph{Proceedings of the IEEE/CVF International Conference on Computer Vision (ICCV)}, pages 4195--4205, 2023.

\bibitem[Radford et~al.(2019)Radford, Wu, Child, Luan, Amodei, and Sutskever]{modelcfg_1}
Alec Radford, Jeffrey Wu, Rewon Child, David Luan, Dario Amodei, and Ilya Sutskever.
\newblock Language models are unsupervised multitask learners.
\newblock \emph{OpenAI blog}, 2019.

\bibitem[Raffel et~al.(2020)Raffel, Shazeer, Roberts, Lee, Narang, Matena, Zhou, Li, and Liu]{rpb_2}
Colin Raffel, Noam Shazeer, Adam Roberts, Katherine Lee, Sharan Narang, Michael Matena, Yanqi Zhou, Wei Li, and Peter~J Liu.
\newblock Exploring the limits of transfer learning with a unified text-to-text transformer.
\newblock \emph{JMLR}, 2020.

\bibitem[Ramesh et~al.(2021)Ramesh, Pavlov, Goh, Gray, Voss, Radford, Chen, and Sutskever]{ar_5}
Aditya Ramesh, Mikhail Pavlov, Gabriel Goh, Scott Gray, Chelsea Voss, Alec Radford, Mark Chen, and Ilya Sutskever.
\newblock Zero-shot text-to-image generation.
\newblock In \emph{Proceedings of the 38th International Conference on Machine Learning}, pages 8821--8831. PMLR, 2021.

\bibitem[Razavi et~al.(2019)Razavi, van~den Oord, and Vinyals]{ar_6}
Ali Razavi, Aaron van~den Oord, and Oriol Vinyals.
\newblock Generating diverse high-fidelity images with vq-vae-2.
\newblock In \emph{Advances in Neural Information Processing Systems}. Curran Associates, Inc., 2019.

\bibitem[Ren et~al.(2024)Ren, Yu, He, Shen, Yuille, and Chen]{flowar}
Sucheng Ren, Qihang Yu, Ju He, Xiaohui Shen, Alan~L. Yuille, and Liang{-}Chieh Chen.
\newblock Flowar: Scale-wise autoregressive image generation meets flow matching.
\newblock \emph{CoRR}, abs/2412.15205, 2024.

\bibitem[Rombach et~al.(2022)Rombach, Blattmann, Lorenz, Esser, and Ommer]{Diff_6}
Robin Rombach, Andreas Blattmann, Dominik Lorenz, Patrick Esser, and Bj\"orn Ommer.
\newblock High-resolution image synthesis with latent diffusion models.
\newblock In \emph{Proceedings of the IEEE/CVF Conference on Computer Vision and Pattern Recognition (CVPR)}, pages 10684--10695, 2022.

\bibitem[Salimans et~al.(2016)Salimans, Goodfellow, Zaremba, Cheung, Radford, and Chen]{IS}
Tim Salimans, Ian Goodfellow, Wojciech Zaremba, Vicki Cheung, Alec Radford, and Xi Chen.
\newblock Improved techniques for training gans.
\newblock In \emph{Proceedings of the 30th International Conference on Neural Information Processing Systems}, page 2234–2242, Red Hook, NY, USA, 2016. Curran Associates Inc.

\bibitem[Shaw et~al.(2018)Shaw, Uszkoreit, and Vaswani]{rpb_3}
Peter Shaw, Jakob Uszkoreit, and Ashish Vaswani.
\newblock Self-attention with relative position representations.
\newblock In \emph{Proceedings of the 2018 Conference of the North American Chapter of the Association for Computational Linguistics: Human Language Technologies, NAACL-HLT, New Orleans, Louisiana, USA, June 1-6, 2018, Volume 2 (Short Papers)}, pages 464--468. Association for Computational Linguistics, 2018.

\bibitem[Shazeer(2020)]{swiglu}
Noam Shazeer.
\newblock {GLU} variants improve transformer.
\newblock \emph{CoRR}, abs/2002.05202, 2020.

\bibitem[Shi et~al.(2024)Shi, Luo, Ge, Yang, Shan, and Wang]{ibq}
Fengyuan Shi, Zhuoyan Luo, Yixiao Ge, Yujiu Yang, Ying Shan, and Limin Wang.
\newblock Taming scalable visual tokenizer for autoregressive image generation.
\newblock \emph{CoRR}, abs/2412.02692, 2024.

\bibitem[Song et~al.(2021)Song, Meng, and Ermon]{Diff_7}
Jiaming Song, Chenlin Meng, and Stefano Ermon.
\newblock Denoising diffusion implicit models.
\newblock In \emph{International Conference on Learning Representations}, 2021.

\bibitem[Su et~al.(2024)Su, Ahmed, Lu, Pan, Bo, and Liu]{1drope}
Jianlin Su, Murtadha H.~M. Ahmed, Yu Lu, Shengfeng Pan, Wen Bo, and Yunfeng Liu.
\newblock Roformer: Enhanced transformer with rotary position embedding.
\newblock \emph{Neurocomputing}, 568:\penalty0 127063, 2024.

\bibitem[Sun et~al.(2024)Sun, Jiang, Chen, Zhang, Peng, Luo, and Yuan]{llamagen}
Peize Sun, Yi Jiang, Shoufa Chen, Shilong Zhang, Bingyue Peng, Ping Luo, and Zehuan Yuan.
\newblock Autoregressive model beats diffusion: Llama for scalable image generation.
\newblock \emph{arXiv preprint arXiv:2406.06525}, 2024.

\bibitem[Team(2024{\natexlab{a}})]{vlm_9}
Chameleon Team.
\newblock Chameleon: Mixed-modal early-fusion foundation models.
\newblock \emph{arXiv preprint arXiv:2405.09818}, 2024{\natexlab{a}}.

\bibitem[Team(2024{\natexlab{b}})]{emu3}
Emu3 Team.
\newblock Emu3: Next-token prediction is all you need.
\newblock \emph{Tech Report}, 2024{\natexlab{b}}.

\bibitem[Team et~al.(2023)Team, Anil, Borgeaud, Wu, Alayrac, Yu, Soricut, Schalkwyk, Dai, Hauth, et~al.]{llm_4}
Gemini Team, Rohan Anil, Sebastian Borgeaud, Yonghui Wu, Jean-Baptiste Alayrac, Jiahui Yu, Radu Soricut, Johan Schalkwyk, Andrew~M Dai, Anja Hauth, et~al.
\newblock Gemini: a family of highly capable multimodal models.
\newblock \emph{arXiv preprint arXiv:2312.11805}, 2023.

\bibitem[Tian et~al.(2024)Tian, Jiang, Yuan, PENG, and Wang]{var}
Keyu Tian, Yi Jiang, Zehuan Yuan, BINGYUE PENG, and Liwei Wang.
\newblock Visual autoregressive modeling: Scalable image generation via next-scale prediction.
\newblock In \emph{The Thirty-eighth Annual Conference on Neural Information Processing Systems}, 2024.

\bibitem[Touvron et~al.(2023{\natexlab{a}})Touvron, Lavril, Izacard, Martinet, Lachaux, Lacroix, Rozi{\`e}re, Goyal, Hambro, Azhar, et~al.]{llm_1}
Hugo Touvron, Thibaut Lavril, Gautier Izacard, Xavier Martinet, Marie-Anne Lachaux, Timoth{\'e}e Lacroix, Baptiste Rozi{\`e}re, Naman Goyal, Eric Hambro, Faisal Azhar, et~al.
\newblock Llama: Open and efficient foundation language models.
\newblock \emph{arXiv preprint arXiv:2302.13971}, 2023{\natexlab{a}}.

\bibitem[Touvron et~al.(2023{\natexlab{b}})Touvron, Martin, Stone, Albert, Almahairi, Babaei, Bashlykov, Batra, Bhargava, Bhosale, et~al.]{llm_2}
Hugo Touvron, Louis Martin, Kevin Stone, Peter Albert, Amjad Almahairi, Yasmine Babaei, Nikolay Bashlykov, Soumya Batra, Prajjwal Bhargava, Shruti Bhosale, et~al.
\newblock Llama 2: Open foundation and fine-tuned chat models.
\newblock \emph{arXiv preprint arXiv:2307.09288}, 2023{\natexlab{b}}.

\bibitem[van~den Oord et~al.(2017)van~den Oord, Vinyals, and kavukcuoglu]{ar_7}
Aaron van~den Oord, Oriol Vinyals, and koray kavukcuoglu.
\newblock Neural discrete representation learning.
\newblock In \emph{Advances in Neural Information Processing Systems}. Curran Associates, Inc., 2017.

\bibitem[Wang et~al.(2024)Wang, Bai, Tan, Wang, Fan, Bai, Chen, Liu, Wang, Ge, Fan, Dang, Du, Ren, Men, Liu, Zhou, Zhou, and Lin]{vlm_5}
Peng Wang, Shuai Bai, Sinan Tan, Shijie Wang, Zhihao Fan, Jinze Bai, Keqin Chen, Xuejing Liu, Jialin Wang, Wenbin Ge, Yang Fan, Kai Dang, Mengfei Du, Xuancheng Ren, Rui Men, Dayiheng Liu, Chang Zhou, Jingren Zhou, and Junyang Lin.
\newblock Qwen2-vl: Enhancing vision-language model's perception of the world at any resolution.
\newblock \emph{CoRR}, abs/2409.12191, 2024.

\bibitem[Weber et~al.(2024)Weber, Yu, Yu, Deng, Shen, Cremers, and Chen]{mask_3}
Mark Weber, Lijun Yu, Qihang Yu, Xueqing Deng, Xiaohui Shen, Daniel Cremers, and Liang-Chieh Chen.
\newblock Maskbit: Embedding-free image generation via bit tokens.
\newblock \emph{arXiv preprint arXiv:2409.16211}, 2024.

\bibitem[Wu et~al.(2024)Wu, Chen, Wu, Ma, Liu, Pan, Liu, Xie, Yu, Ruan, and Luo]{vlm_10}
Chengyue Wu, Xiaokang Chen, Zhiyu Wu, Yiyang Ma, Xingchao Liu, Zizheng Pan, Wen Liu, Zhenda Xie, Xingkai Yu, Chong Ruan, and Ping Luo.
\newblock Janus: Decoupling visual encoding for unified multimodal understanding and generation.
\newblock \emph{CoRR}, abs/2410.13848, 2024.

\bibitem[Yu et~al.(2022)Yu, Li, Koh, Zhang, Pang, Qin, Ku, Xu, Baldridge, and Wu]{ar_8}
Jiahui Yu, Xin Li, Jing~Yu Koh, Han Zhang, Ruoming Pang, James Qin, Alexander Ku, Yuanzhong Xu, Jason Baldridge, and Yonghui Wu.
\newblock Vector-quantized image modeling with improved {VQGAN}.
\newblock In \emph{International Conference on Learning Representations}, 2022.

\bibitem[Yu et~al.(2023)Yu, Cheng, Sohn, Lezama, Zhang, Chang, Hauptmann, Yang, Hao, Essa, and Jiang]{mask_4}
Lijun Yu, Yong Cheng, Kihyuk Sohn, Jos\'e Lezama, Han Zhang, Huiwen Chang, Alexander~G. Hauptmann, Ming-Hsuan Yang, Yuan Hao, Irfan Essa, and Lu Jiang.
\newblock Magvit: Masked generative video transformer.
\newblock In \emph{Proceedings of the IEEE/CVF Conference on Computer Vision and Pattern Recognition (CVPR)}, pages 10459--10469, 2023.

\bibitem[Yu et~al.(2024{\natexlab{a}})Yu, Lezama, Gundavarapu, Versari, Sohn, Minnen, Cheng, Gupta, Gu, Hauptmann, et~al.]{mask_5}
Lijun Yu, Jos{\'e} Lezama, Nitesh~B Gundavarapu, Luca Versari, Kihyuk Sohn, David Minnen, Yong Cheng, Agrim Gupta, Xiuye Gu, Alexander~G Hauptmann, et~al.
\newblock Language model beats diffusion--tokenizer is key to visual generation.
\newblock In \emph{ICLR}, 2024{\natexlab{a}}.

\bibitem[Yu et~al.(2024{\natexlab{b}})Yu, He, Deng, Shen, and Chen]{rar}
Qihang Yu, Ju He, Xueqing Deng, Xiaohui Shen, and Liang-Chieh Chen.
\newblock Randomized autoregressive visual generation.
\newblock \emph{arXiv preprint arXiv:2411.00776}, 2024{\natexlab{b}}.

\bibitem[Yu et~al.(2024{\natexlab{c}})Yu, Weber, Deng, Shen, Cremers, and Chen]{titok}
Qihang Yu, Mark Weber, Xueqing Deng, Xiaohui Shen, Daniel Cremers, and Liang-Chieh Chen.
\newblock An image is worth 32 tokens for reconstruction and generation.
\newblock In \emph{The Thirty-eighth Annual Conference on Neural Information Processing Systems}, 2024{\natexlab{c}}.

\bibitem[Zhang et~al.(2022)Zhang, Roller, Goyal, Artetxe, Chen, Chen, Dewan, Diab, Li, Lin, Mihaylov, Ott, Shleifer, Shuster, Simig, Koura, Sridhar, Wang, and Zettlemoyer]{rmsnorm}
Susan Zhang, Stephen Roller, Naman Goyal, Mikel Artetxe, Moya Chen, Shuohui Chen, Christopher Dewan, Mona~T. Diab, Xian Li, Xi~Victoria Lin, Todor Mihaylov, Myle Ott, Sam Shleifer, Kurt Shuster, Daniel Simig, Punit~Singh Koura, Anjali Sridhar, Tianlu Wang, and Luke Zettlemoyer.
\newblock {OPT:} open pre-trained transformer language models.
\newblock \emph{CoRR}, abs/2205.01068, 2022.

\bibitem[Zhao et~al.(2024)Zhao, Song, Wang, Feng, Ding, Sun, Xiao, and Wang]{mono}
Chuyang Zhao, Yuxing Song, Wenhao Wang, Haocheng Feng, Errui Ding, Yifan Sun, Xinyan Xiao, and Jingdong Wang.
\newblock Monoformer: One transformer for both diffusion and autoregression.
\newblock \emph{arXiv preprint arXiv:2409.16280}, 2024.

\bibitem[Zhu et~al.(2023)Zhu, Chen, Shen, Li, and Elhoseiny]{vlm_2}
Deyao Zhu, Jun Chen, Xiaoqian Shen, Xiang Li, and Mohamed Elhoseiny.
\newblock Minigpt-4: Enhancing vision-language understanding with advanced large language models.
\newblock \emph{arXiv preprint arXiv:2304.10592}, 2023.

\end{thebibliography}
